\documentclass[journal= iecred]{achemso} 
\setkeys{acs}{doi=true}
\usepackage[inkscapeformat=png]{svg}
\usepackage[super,sort&compress]{natbib} 
\usepackage{rotating}    
\usepackage{listings}
\usepackage{caption}
\usepackage{ragged2e} 
\usepackage{chngcntr}
\usepackage{enumitem}
\usepackage{soul}
\usepackage{booktabs}
\usepackage{geometry}   
\usepackage{longtable}  
\usepackage{multirow}   
\usepackage{caption}    
\usepackage{siunitx}    
\usepackage{subcaption} 
\usepackage{svg}
\usepackage[T1]{fontenc}
\usepackage[utf8]{inputenc}
\usepackage{enumerate}
\usepackage{babel}
\usepackage[numbers]{natbib} 
\usepackage{csquotes}
\nocite{*}
\usepackage{url}
\usepackage{xurl}
\newcommand\rurl[1]{%
  \href{https://#1}{\nolinkurl{#1}}%
}
\usepackage{xr}    
\usepackage[colorlinks=true,        
            linkcolor=blue,
            citecolor=blue,
            urlcolor=blue,
            hypertexnames=false]{hyperref}
\usepackage{setspace}
\usepackage{makecell}
\usepackage{pdflscape}
\usepackage[most]{tcolorbox}
\usepackage{adjustbox}
\usepackage{array,tabularx}           
\usepackage{titlesec}
\usepackage[version=4]{mhchem}    
\usepackage{amsmath,amssymb}      
\usepackage{xcolor}
\lstdefinelanguage{YAML}{
  basicstyle=\ttfamily\small,
  showstringspaces=false,
  breaklines=true,
  alsoletter={-},
  morecomment=[l]{\#},
  moredelim=**[is][\color{blue!60!black}]{@}{@}, 
  literate =
    {---}{{\textemdash}}3
    {~}{{\textasciitilde}}1
}

\lstdefinestyle{ymlstyle}{
  language=YAML,
  frame=single,
  numbers=left,
  numberstyle=\tiny,
  numbersep=6pt,
  columns=fullflexible,
  postbreak=\mbox{\textellipsis\space}, 
  tabsize=2,
  keywordstyle=\color{ymlKey}\bfseries,
  commentstyle=\color{ymlComment}\itshape,
  stringstyle=\color{ymlString},
  backgroundcolor=\color{black!2},
  rulecolor=\color{black!25},
}

\title{Batch Distillation Data for Developing Machine Learning Anomaly Detection Methods}
\author{Justus Arweiler}
\author{Indra Jungjohann}
\affiliation[LTD]
{Laboratory of Engineering Thermodynamics, RPTU Kaiserslautern, Erwin-Schrödinger-Straße 44, 67663 Kaiserslautern, Germany}
\author{Aparna Muraleedharan}
\affiliation[TUM]
{Technical University of Munich, Campus Straubing for Biotechnology and Sustainability, Laboratory for Chemical Process Engineering, Uferstraße 53, 94315 Straubing, Germany}
\author{Heike Leitte}
\affiliation[CS]
{Department of Computer Science, RPTU Kaiserslautern, Erwin-Schrödinger-Straße 44, 67663 Kaiserslautern, Germany}
\author{Jakob Burger}
\affiliation[TUM]
{Technical University of Munich, Campus Straubing for Biotechnology and Sustainability, Laboratory for Chemical Process Engineering, Uferstraße 53, 94315 Straubing, Germany}
\author{Kerstin Münnemann}
\author{Fabian Jirasek}
\email{fabian.jirasek@rptu.de}
\author{Hans Hasse}
\affiliation[LTD]
{Laboratory of Engineering Thermodynamics, RPTU Kaiserslautern, Erwin-Schrödinger-Straße 44, 67663 Kaiserslautern, Germany} 

\begin{document}
\begin{abstract}
    Machine learning (ML) holds great potential to advance anomaly detection (AD) in chemical processes. However, the development of ML-based methods is hindered by the lack of openly available experimental data. To address this gap, we have set up a laboratory-scale batch distillation plant and operated it to generate an extensive experimental dataset, covering fault-free experiments and experiments in which anomalies were intentionally induced, for training advanced ML-based AD methods. In total, 119 experiments were conducted across a wide range of operating conditions and mixtures. Most experiments containing anomalies were paired with a corresponding fault-free one. The dataset that we provide here includes time-series data from numerous sensors and actuators, along with estimates of measurement uncertainty. In addition, unconventional data sources -- such as concentration profiles obtained via online benchtop NMR spectroscopy and video and audio recordings -- are provided. Extensive metadata and expert annotations of all experiments are included. The anomaly annotations are based on an ontology developed in this work. The data are organized in a structured dataset and made freely available via \rurl{doi.org/10.5281/zenodo.17395543}. This new dataset paves the way for the development of advanced ML-based AD methods. As it includes information on the causes of anomalies, it further enables the development of interpretable and explainable ML approaches, as well as methods for anomaly mitigation.
\end{abstract}

\newpage
\section{Background \& Summary}
\label{sec:Introduction}
In chemical process engineering, early and reliable anomaly detection (AD) is highly relevant for preventing costly plant failures and malfunctions~\cite{Chandola2009}. Undetected anomalies in chemical processes can pose serious hazards, threatening human health and the environment. They can also cause significant economic losses. To reduce such risks, anomalies must be promptly and reliably identified and mitigated through appropriate anomaly management (AM). The large amount of diverse data generated in chemical processes makes automated AD mandatory. AD methods are therefore implemented in virtually all industrial process control systems. Nevertheless, human operators still play a major role in both AD and AM.

In principle, AD can be considered as a classification task, where, in the simplest case, only two classes are distinguished: normal and anormal. The design of AD methods is not trivial, as it is characterized by conflicting objectives: good scores in true positive (anomaly detected when it occurs) tend to go hand in hand with bad scores in false positive (anomaly detected in normal operation), and good scores in true negative (no anomaly detected in normal operation) tend to go hand in hand with bad scores in false negative (no anomaly detected even though present).
A wide variety of AD methods have been described in the literature. The proposed approaches differ so significantly that even their classification is challenging~\cite {Venkatasubramanian2003}. A coarse classification distinguishes between methods that rely only on process data (data-driven methods) and others that combine these data with a process model (model-based methods). For an overview of existing classical AD methods, we refer the reader to reviews of the topic, e.g.,~\cite{Chandola2009, Venkatasubramanian2003, Venkatasubramanian2003a, Venkatasubramanian2003b, Chiang2001, Hodge2004}.

The rapid development of machine learning (ML) has opened up new opportunities in many fields, including chemical engineering, that are being explored by many researchers~\cite{Mowbray2022, Dobbelaere2021, Goettl2025, Gond2025, Hayer2022, Hayer2025, Hayer2025a, Hayer2025b, Hoffmann2025, Jirasek2023, Jirasek2020, Jirasek2021, Specht2023, Specht2024, Vollmer2024}. It is, therefore, astonishing that only comparatively few papers exist in which advanced ML methods, most notably deep learning methods, have been applied to AD in chemical processes~\cite{Chadha2019, Monroy2009, Inoue2017, Song2019, Tian2020, Wu2024, Hartung2023}. This gap can only partially be explained by the fact that AD on high-dimensional complex time-series data has only recently attracted wider attention in the ML community, as this field of ML has developed rapidly and yielded a large number of methods that could principally be applied to AD in chemical processes~\cite{Hartung2023, Russell2000, Schmidl2022, Darban2024, Wang2025}. The main problem that hinders the application of modern ML methods to AD in chemical processes is the lack of suitable data for training, validation, and testing. Although such data are abundant in chemical companies, they are not available to academic researchers. This shortage of accessible process data significantly hinders the development of advanced ML-based AD techniques, which rely heavily on large datasets for training. In contrast, classical AD methods, which primarily use data for testing, are less affected by this limitation~\cite{Venkatasubramanian2003}.

Up until now, the only larger dataset available for the development, training, and testing of ML-based AD methods for chemical processes has been the Tennessee-Eastman Process (TEP) dataset~\cite{Downs1993, Rieth2017, Hartung2023}. The TEP represents a fictitious continuous chemical process composed of five key unit operations: a reactor, a product condenser, a vapor-liquid separator, a recycle compressor, and a product stripper. As the TEP data were generated through simulation, they differ significantly from real-world process data. Although some noise was artificially superimposed, the data are still, in essence, deterministic as they are based on a physical process model that reflects only some features of real processes. As a result, the data are considerably better behaved than real process data. We have recently applied advanced ML-based AD methods on the TEP data and have obtained excellent results~\cite{Hartung2023}. However, preliminary tests using data from the plant described in the present paper, applying the same AD methods, have yielded only very poor results. This stark contrast highlights the critical need for real process data for the development and evaluation of ML-based AD techniques. Furthermore, in the ML literature, deep learning architectures, such as Long Short-Term Memory (LSTM) networks, convolutional neural networks (CNNs), autoencoders, and, more recently, transformer-based models, have been explored for AD in time-series data. While these approaches often show strong performance on synthetic or benchmark datasets, their effectiveness remains largely unvalidated on real industrial process data~\cite{Chadha2019, Monroy2009, Inoue2017, Song2019, Tian2020, Wu2024, Hartung2023}.

Therefore, we provide such data in the present work. An obviously crucial question is thereby: can suitable data be obtained in an academic environment? We argue that this is possible and may even offer advantages over data from full-scale production sites. Most notably, academic environments provide greater flexibility in designing and executing experimental scenarios, and typically allow for more comprehensive documentation. The most basic decision in planning such experiments is choosing a suitable demonstration process. We chose distillation for three reasons: first, it is a very widely applied separation process in the chemical industry; second, it lends itself to scale-up: we can operate plants on small scales (as we have to in the academic environment) and still learn from them about the behavior of full-scale production plants; and third, there are excellent physical models for distillation, which can be leveraged for data augmentation and the creation of hybrid datasets for ML model training in future work. 

Chemical production processes are commonly classified as either batch or continuous. We report here on batch distillation. The studied batch distillation process is inherently unsteady. Complementary data from a steady-state continuous distillation process are currently being collected at TU Munich, within the same coordinated project (see~\hyperref[sec:BATCH_Acknowledgements]{Funding}, DFG Research Unit 5359 “Deep Learning on Sparse Chemical Process Data”), on which we report in a separate paper~\cite{APARNA, AparnaDataset}. These data will be released in a format consistent with the present dataset. 

Although the batch distillation plant described here is a small, flexible laboratory plant, it can emulate most types of faults that occur in industrial-scale operations~\cite{Kister2003}. In this paper, we describe the batch distillation plant and its operation in sufficient detail to contextualize the generated process data. The plant was specifically designed and equipped for the present study, and operated under a wide range of conditions and mixtures, including both fault-free and anomalous experiments. Alongside conventional process variables, the dataset includes multimodal information, such as concentration data from online NMR spectroscopy, as well as video and audio recordings. We also provide information on the types of anomalies that were generated. Expert annotations describe the anomalies and their causes based on an ontology developed in this work.

With this paper, we present the first extensive dataset of real batch distillation process data. This paper establishes the foundation for follow-up publications in which this dataset will be enlarged and diversified with new process data generated with the described batch distillation plant. Future work will also extend the dataset to additional operating modes of the batch distillation plant, including dynamic control and fed-batch operation. We openly share our data with the community to foster progress in ML-based AD in chemical engineering; the complete dataset is publicly available in a structured format under a CC BY 4.0 license at \rurl{doi.org/10.5281/zenodo.17395543}\cite{MainDataset}.  

\section{Methods}
\subsection{Batch Distillation Plant}
\subsubsection{Overview}
\label{sec:BATCH_Plant}
\begin{figure}[H]
    \centering
    \includegraphics[width=1.0\textwidth]{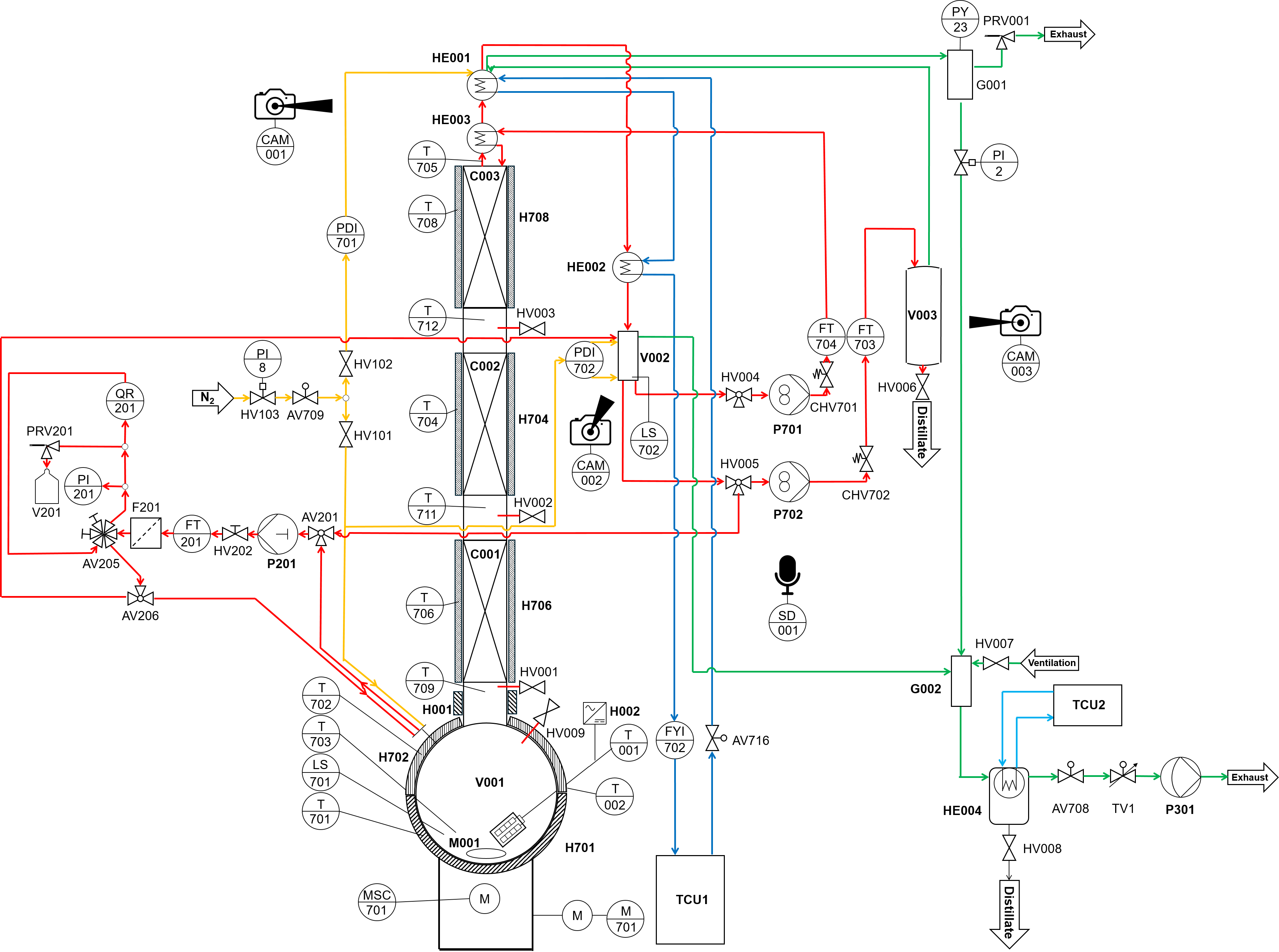}
    \captionsetup{justification=justified, singlelinecheck=false} 
    \caption{P\&I diagram of the batch distillation plant. Color code of the lines: product (red), cooling water (dark blue), cooling ethanol (light blue), nitrogen (yellow), and pressure control (green). The equipment list, in which the labels are declared, is given in the Supplementary Information.}
    \label{fig:PID_Batch}
\end{figure}

\begin{figure}[H]
    \centering
    \includegraphics[width=0.9\textwidth, keepaspectratio]{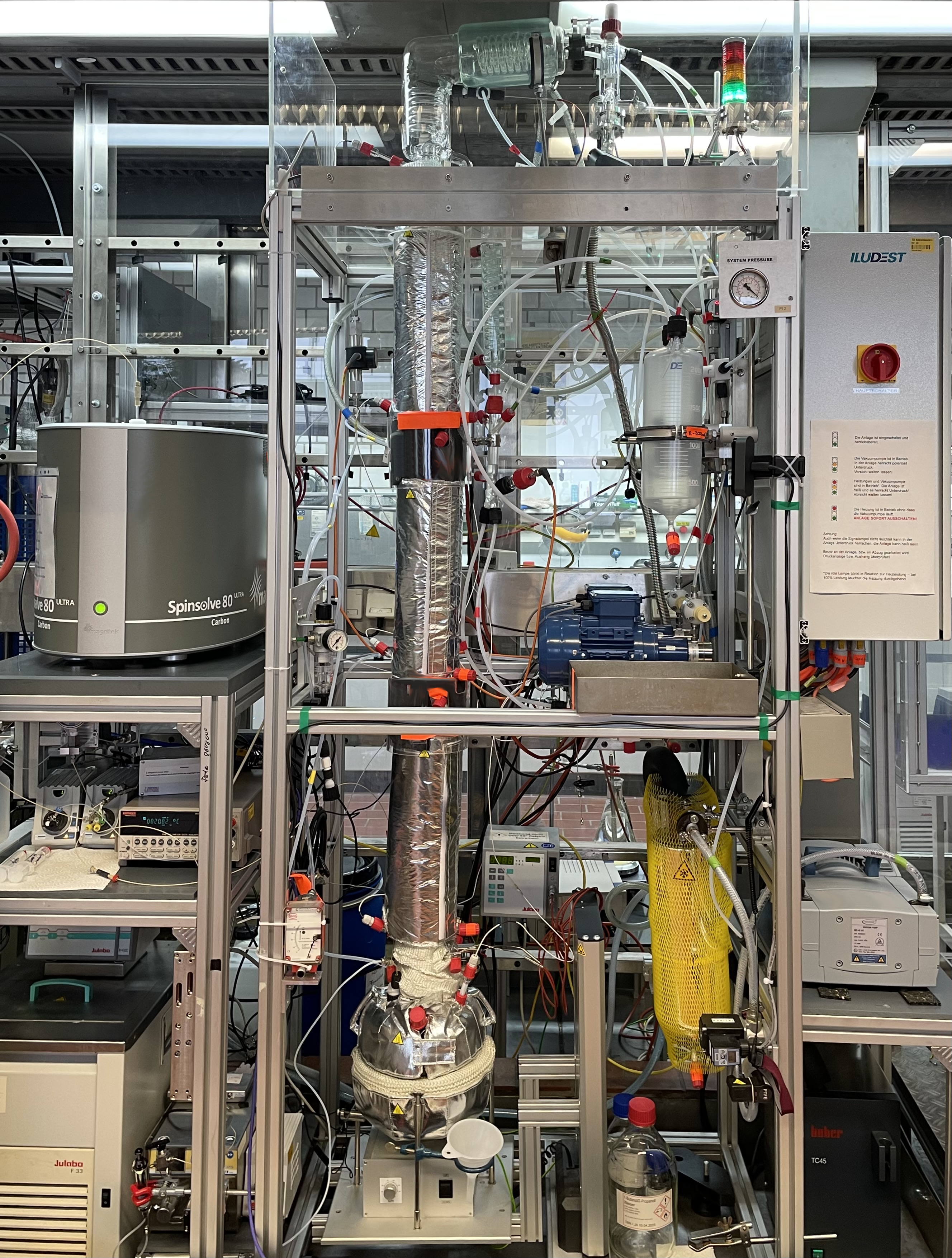}
    \captionsetup{justification=justified, singlelinecheck=false}
    \caption{Photo of the laboratory batch distillation plant. The reboiler vessel (V001) and the distillation column (C001-C003) are surrounded by insulation. The visible glass apparatuses are the condenser (HE001-HE003), the buffer vessel (V002), and the distillate receiver (V003). The pumps (P701, P702) are blue, the condensation trap (HE004) has a yellow jacket and is connected to the gray vacuum pump (P301). The benchtop NMR spectrometer is on the top left rack.}
    \label{fig:Picture_Batch}
\end{figure}

The laboratory-scale batch distillation plant used to generate dynamic process data is a modified version of the Iludest LM 2/S glass plant. Figures~\ref{fig:PID_Batch} and~\ref{fig:Picture_Batch} show the P\&I diagram and a photograph of the plant, respectively. A complete list of the equipment, measurement instrumentation, and controls, including the respective suppliers, is provided in the Supplementary Information. It also includes a mapping to the labels in the P\&I diagram of Figure~\ref{fig:PID_Batch} which are used in the discussion below. The different types of sensors and control equipment are classified here according to the Sensor, Observation, Sample, and Actuator (SOSA) ontology~\cite{sosa}. 

The setup comprises a stirred 2 L reboiler vessel (V001) with a distillation glass column (inner diameter 30 mm) on top. The column has three sections (C001-C003), each equipped with structured laboratory packing (Sulzer DX30, 37.5 cm height), corresponding to approximately 12 theoretical stages for the entire column. The column segments are thermally insulated by an evacuated mirrored double jacket. Heat is supplied to the reboiler vessel V001 by a heating jacket (H701, 350 W) and a submerged electric heating element (H002, 400 W). The column and the reboiler vessel are jacketed, with heating elements (H702-H708) wrapped in thermal insulation to prevent heat loss through the jackets. The overhead vapor is condensed and subcooled in glass heat exchangers (HE001, HE002) and then collected in a buffer vessel (V002). Cooling water is supplied by a thermostat (TCU1) at~283.15~K. Two gear-type pumps (P701, P702) deliver the condensate either to the distillate receiver (V003) or, as reflux, back into the top of the column. Before the subcooled reflux is returned to the column, it is warmed up in the condenser (HE003). The pressure in the plant is subambient and controlled using a membrane vacuum pump (P301). To protect the vacuum pump from solvent residues, a condensation trap (HE004) is used, operated with liquid ethanol at~233.15~K supplied by a thermostat (TCU2). The plant can be purged with nitrogen. In the experiments of the present work, the plant was operated at fixed pressure, reflux ratio $R$, and heat supply. The reflux ratio $R$ is the ratio of the mass flow rates of the distillate pumped back into the distillation column, $\dot{m}_{\text{FT704}}$, and that pumped into the distillate receiver V003, $\dot{m}_{\text{FT703}}$:
\begin{equation}
    R = \frac{\dot{m}_{\text{FT704}}}{\dot{m}_{\text{FT703}}}
\end{equation}
As a constant liquid level in the buffer vessel V002 is maintained, the sum of $\dot{m}_{\text{FT703}}$ and $\dot{m}_{\text{FT704}}$ is basically equal to the total mass flow of the condensate (neglecting the small losses into the condensation trap HE004).

\subsubsection{Actuators}
\label{subsec:BATCH_Actuators}
Flexible operation of the batch distillation plant is achieved through the use of diverse actuators, including manually operated components and electronically controlled elements. These actuators serve various purposes, such as controlling fluid flow, heat supply, and pressure. Manually operated actuators, such as sampling valves and ventilation ports, are primarily used for system maintenance and process interventions that require human oversight. In contrast, the actuators integrated into the process control system (PCS, cf. Section~\nameref{subsec:BATCH_Control}) enable automated, precise, and reproducible operation of the plant. A list of actuators that can be controlled through the PCS is given in Table~\ref{tab:BATCH_actuators}. 
\begin{table}[H]
    \centering
    \captionsetup{justification=raggedright, singlelinecheck=false}
    \caption{List of actuators in the batch distillation plant that are integrated in the process control system. The labels are the same as in the P\&I diagram in Figure~\ref{fig:PID_Batch}.}
   {%
    \begin{tabular}{l l l l l}
    \hline
        \makecell[tl]{Equipment\\type} & Label & \makecell[tl]{Controlled\\variable} & Unit & Description  \\ \hline \hline
        Heater & H002 & Power& \% &  \makecell[tl]{Power supply to the heating\\rod submerged in reboiler vessel V001}\\
                      & H701 & Power & \% &  \makecell[tl]{Power supply to the heating\\jacket of reboiler vessel V001}\\
                      & H702 & Power & \% &  \makecell[tl]{Power supply to the heating\\jacket of reboiler vessel V001}\\
                      & H704 & Power &\% &  \makecell[tl]{Power supply to the heating\\jacket of column section C001}\\
                      & H706 &  Power &\% &  \makecell[tl]{Power supply to the heating\\jacket of column section C002}\\
                      & H708 & Power &\% &  \makecell[tl]{Power supply to the heating\\jacket of column section C003}\\
      \addlinespace[4pt]
        Pump  & P201 & Flow rate & ml min$^{-1}$ & \makecell[tl]{Volumetric flow rate of liquid \\ to online NMR}\\
              & P301 & State & on/off & \makecell[tl]{Vacuum pump\\for system pressure regulation}\\
              & P701 & Power& \% & Power supply of reflux pump \\
              & P702 & Power& \% & \makecell[tl]{Power supply of distillate\\withdrawal pump}\\
      \addlinespace[4pt]
        Valve  & AV708 & State & open/closed & State of valve at vacuum pump line \\
               & AV716 & State& open/closed & State of valve at cooling water line \\
               & TV1 & \makecell[tl]{Open cross\\section}& \% & \makecell[tl]{Open cross section of the vacuum\\throttle valve}\\
    \hline
    \end{tabular}
    }
    \label{tab:BATCH_actuators}
\end{table}

\subsubsection{Sensors}
\label{subsec:BATCH_Sensors}
All conventional sensors of the plant are listed in Table~\ref{tab:BATCH_sensors}. Table~\ref{tab:BATCH_sensorsuppliers} gives an overview of the used sensor types and their respective measurement uncertainty. The temperatures of the reboiler vessel, the condenser, and the three column sections, as well as the temperatures of each electrical heating element, are tracked. Moreover, flow rates of the product, reflux, and cooling water, and the levels of the liquid in the reboiler vessel V001 and the buffer vessel V002 are tracked. The pressure in the condenser and the differential pressure between the bottom and the top of the column are monitored. Information on sensor calibration is provided in the Supplementary Information.

\begin{table}[H]
    \centering
    \captionsetup{justification=raggedright, singlelinecheck=false}
    \caption{List of sensors of the batch distillation plant. The labels are the same as in the P\&I diagram in Figure~\ref{fig:PID_Batch}. Sensors marked with a dagger ($^\dag$) are part of a control loop.}
    \begin{tabular}{l l l l l l l}
    \hline
    \makecell[tl]{Measured\\variable}  & \makecell{Label} &  Unit & Feature of interest \\ \hline\hline
    Level & LS701  &  -  & \makecell[tl]{Contact to liquid (0: dry, 1: wet)\\ in reboiler vessel V001}\\
     & LS702  &  - & \makecell[tl]{Contact to liquid (0: dry, 1: wet)\\ in buffer vessel V002}\\
    \addlinespace[4pt]
    Mass flow & FT201    &  kg h$^{-1}$ & Liquid sample stream to online NMR \\
     & FT703$^\dag$  &  kg h$^{-1}$ & Top product stream \\
     & FT704$^\dag$ &  kg h$^{-1}$ & Reflux stream \\ 
     & FYI702 &  s$^{-1}$ & Cooling water stream\\
    \addlinespace[4pt]
    \makecell[tl]{Pressure\\ difference} & PDI701  &  mbar  & \makecell[tl]{Pressure drop along column\\ (reboiler vessel V001 to top vessel G001)}\\
    & PDI702$^\dag$  & mbar  & Static pressure in buffer vessel V002\\
    \addlinespace[4pt]
    Pressure   & PY23$^\dag$  &    mbar & Pressure in top vessel G001\\
    \addlinespace[4pt]
    Temperature & T701 &  °C & \makecell[tl]{Heating jacket H701 for lower hemisphere\\of reboiler vessel V001}\\
    & T702$^\dag$    &  °C & \makecell[tl]{Heating jacket H701 for upper hemisphere\\of reboiler vessel V001}\\
    & T703$^\dag$ &  °C & Liquid residue in reboiler vessel V001 \\
    & T704$^\dag$  &  °C & Heating jacket H704 \\
    & T705$^\dag$ &  °C & Fluid on top of column section C003 \\
    & T706$^\dag$ &  °C & Heating jacket H706 \\
    & T708$^\dag$ &  °C & Heating jacket H708 \\
    & T709 &  °C & Fluid beneath column section C001 \\
    & T711 &  °C & Fluid between column sections C001 and C002 \\
    & T712  &  °C & Fluid between column sections C002 and C003 \\
    \hline  
    \end{tabular}
    \label{tab:BATCH_sensors}
\end{table}

\begin{table}[H]
    \centering
    \captionsetup{justification=raggedright, singlelinecheck=false}
    \caption{Sensor types and measurement uncertainties. The labels are those of the corresponding components in the P\&I diagram in Figure~\ref{fig:PID_Batch}. The actual measurement uncertainty may depend on the measured value.}    \resizebox{\textwidth}{!}{%
    \begin{tabular}{l l l l}
    \hline
        \makecell{Measured \\ variable} & Labels & Sensor type & Uncertainty \\
    \hline \hline
        Level & LS701, LS702 & Diffuse-reflective photoelectric sensor & n.a. \\
        \addlinespace[4pt]
        Mass flow   & FT703, FT704 & Coriolis mass flow sensor & $\pm 0.06 \%$\\
        & FT201 & Coriolis mass flow sensor & $\pm 0.2 \%$ \\ 
        & FYI702 & Paddlewheel flow sensor & $\pm 15\%$ \\
        \addlinespace[4pt]
        \makecell[tl]{Pressure\\ difference} & PDI701, PDI702 & Piezoresistive sensor & $\pm 1 \%$ \\
       \addlinespace[4pt]
        Pressure & PY23 & Capacitance manometer sensor & $\pm 0.05 \%$ \\
        \addlinespace[4pt]
        Temperature  &   T701, T702, T704, & Pt100 class A &  $\pm 0.95$ K\\
         &  T706, T708 & & \\
        & T703, T705, T709, & Pt100 class 1/3 & $\pm 0.1 $ K\\
        & T711, T712 & & \\
    \hline
    \end{tabular}
    }
    \label{tab:BATCH_sensorsuppliers}
\end{table}

\begin{figure}[H]
    \centering
    \includegraphics[width=\linewidth]{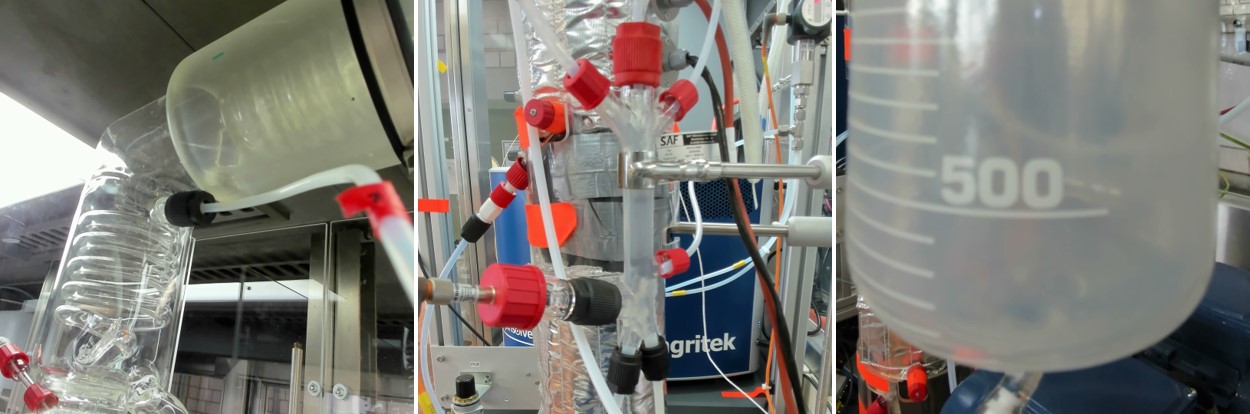}
    \captionsetup{justification=justified, singlelinecheck=false}
    \caption{Snapshots from the cameras in the batch distillation plant. Left: Glass heat exchangers HE001 and HE003. Middle: Buffer vessel V002 with connections. Right: Distillate receiver V003.}
    \label{fig:Batch_CameraImages}
\end{figure}

In addition to the conventional sensors described above, the plant is equipped with several unconventional sensors. Three Logitech BRIO Ultra-HD Pro Business webcams are placed at different locations within the plant to capture high-resolution images every 2 seconds. Sample images are shown in Figure~\ref{fig:Batch_CameraImages}. One camera (CAM001) monitors the glass condensers HE001 and HE003, including the condensate drain and the reflux inlet. The two other cameras monitor the filling levels of the buffer vessel V002 (CAM002) and the distillate receiver V003 (CAM003). Furthermore, audio signals from plant operations are collected with a RØDE AI-Micro microphone, which was either placed beneath the reboiler vessel V001 or between the reflux and distillate withdrawal pumps P701 and P702.

Online concentration measurements were carried out with a Spinsolve 80 ULTRA Carbon benchtop NMR spectrometer (QR201) from Magritek operated with the Magritek software “Spinsolve” Version 2.3.8. Liquid from either the reboiler vessel V001 or the buffer vessel V002 was continuously pumped through a flow cell in the NMR spectrometer using pump P201. Details on the online NMR measurements and the data analysis are given in the Supplementary Information.

In addition to the online NMR analysis, samples were also analyzed offline by gas chromatography. Liquid samples were drawn from the reboiler vessel V001 and the buffer vessel V002 using a gas-tight syringe to prevent air from entering the plant during sampling. The samples were collected approximately every 20 minutes during the experimental runs. Further details on the offline concentration analysis are given in the Supplementary Information. 

\subsubsection{Process Control}
\label{subsec:BATCH_Control}
The batch distillation plant is operated using a Python-based process control system (PCS) implemented in the itom software package~\cite{Gronle2014}, version 4.2.2, an open-source integrated development environment (IDE) for automation. Communication between the PCS and the batch distillation plant is realized using the OPC/UA data exchange format, with software from Iludest called “IluLab” acting as an intermediate communicator. The PCS directly communicates with the electronic lab journal “elabFTW”~\cite{Carpi2017}, which is also used for planning the experimental campaigns and storing the metadata for each experiment. 

Before a distillation experiment is conducted, the corresponding experimental run is created in the electronic lab journal. All sensor data from the PCS are read and stored every second. Furthermore, all events in the PCS, e.g., the start and stop of experiments, changes in actuator settings, violations of security thresholds, or moving to the next step in an experimental recipe, are stored in a log file with corresponding timestamps. Metadata stored for each experiment in the electronic lab journal include ambient conditions, as described in the Supplementary Information, anonymized information about the operators conducting the experimental run, and information on the plant. Changes to the setup or control system, and the operating times of each plant component, are tracked in the electronic lab journal. Hence, the currently installed version of the plant, the feed mixture, the operating point, and information on anomalies are linked to the run. Furthermore, a distillation recipe containing start-up and shut-down instructions for the PCS and operating point settings is linked to the experiment.

For each experiment, the corresponding distillation recipe is loaded into the PCS, enabling standardized procedures for conducting experiments and for plant operation at defined operating points. The operating point of the plant is fixed by set values for the absolute pressure at the column top (PY23), the heat supply to the reboiler vessel (H701 and H002), and the reflux ratio $R$. Thereby, the maximum pressure is atmospheric, and the maximum heat supply is 750 W. The control values for the actuators are set based on the operating point.

Several actuators are operated using PI and PID controllers; detailed information about the controllers is given in Table~\ref{tab:BATCH_controllers}. 
\begin{table}[H]
    \centering
    \captionsetup{justification=justified, singlelinecheck=false}
    \caption{Overview of control circuits used in the batch distillation plant. The labels are those of the components in the P\&I diagram in Figure~\ref{fig:PID_Batch}.}
    \begin{tabular}{ccc}
    \hline
       Actuator &  Target & Type \\ \hline \hline
       H702 & T703 & PI \\
       H704 & T703 & PI\\
       H706 & T703 & PI\\
       H708 & T705 & PI\\
       P701 & FT704, PDI702 & PI \\
       P702 & FT703, PDI702 & PI \\
       TV1  & PY23 & PI \\
       P201 & FT201 & PID\\
       \hline
    \end{tabular}
    \label{tab:BATCH_controllers}
\end{table}

\subsection{Operation}
\label{sec:BATCH_operation}
\subsubsection{General Procedure}
\label{subsec:BATCH_operation_general}
Before each experiment, a leakage test was performed: the dry, empty plant was sealed, and the pressure was set to $p_{\text{PY23}}=10$ mbar. After reaching this pressure, the pressure control was stopped, and the pressure was tracked. The test was considered to be passed if the pressure increase was below 0.5 mbar per hour. 

\begin{figure}[H]
    \centering
    \includegraphics[width=0.95\textwidth]{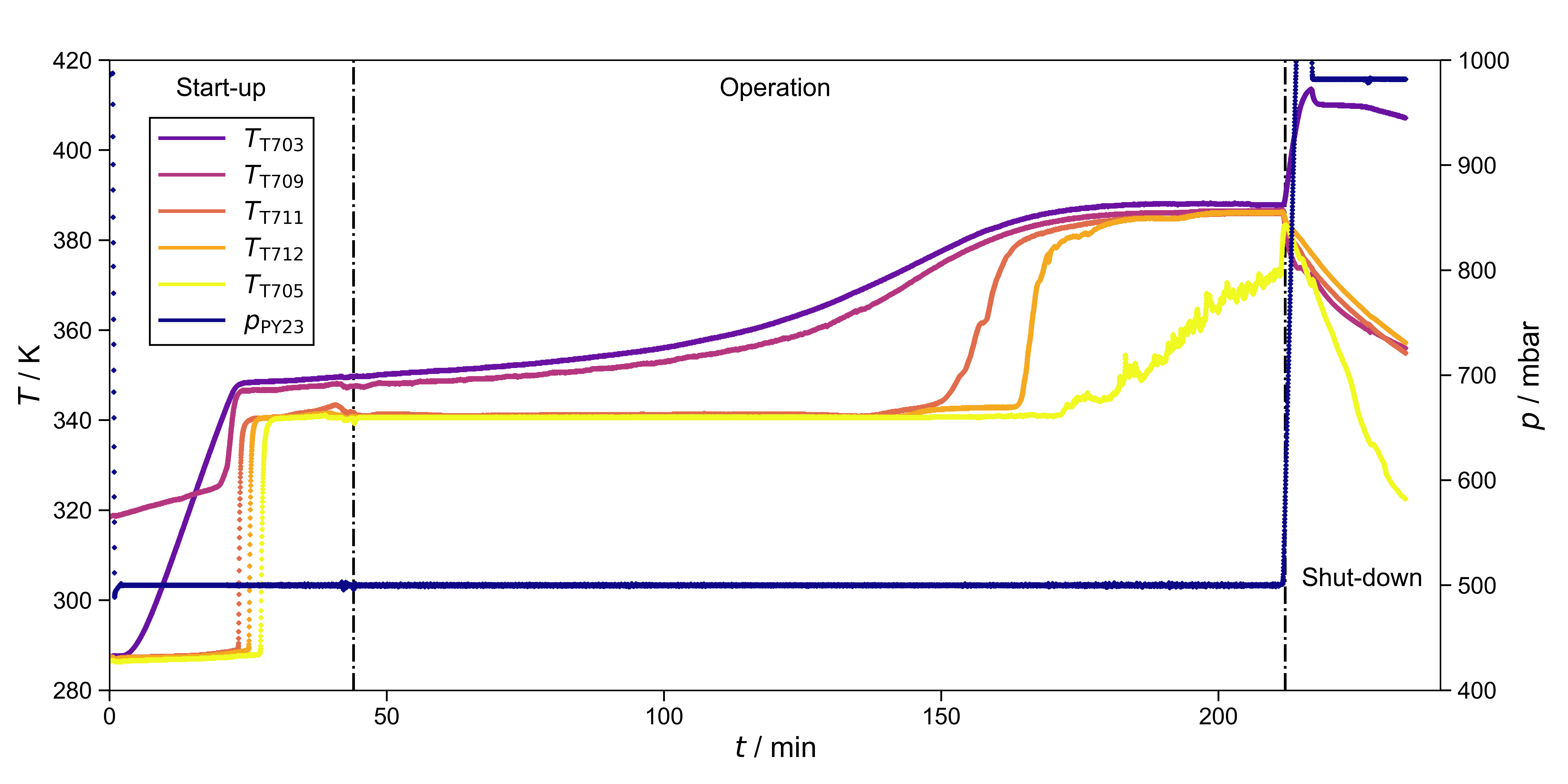}
    \captionsetup{justification=justified, singlelinecheck=false}
    \caption{Temperatures and pressure recorded during experiment \nolinkurl{batch\_dist\_ternary\_butan-1-ol+propan-2-ol+water/operating\_point\_001/test\_anormal\_experiment\_013}. The three phases, start-up, operation, and shut-down are indicated.}
    \label{fig:Batch_TemporalScheme}
\end{figure}

Before the start of the experiment, the plant was flushed with nitrogen, and 1-1.5 kg of the liquid feed mixture of known composition were filled into the reboiler vessel V001. Three phases of the experiment are distinguished here: start-up, operation (either normal or with anomalies), and shut-down; an example is shown in Figure~\ref{fig:Batch_TemporalScheme}. During start-up, first, the desired operating pressure is set. When the set pressure is attained, the primary heat sources, H701 and H002, are ramped up to the specified power. The liquid in the reboiler vessel is first heated up before it begins to boil. The rising vapors cause a rapid increase in temperature in the column sections C001-C003. The overhead vapor condenses in HE001 and HE003. The condensate is collected in the buffer vessel V002 and, during start-up, is entirely returned to the column as liquid reflux. The start-up of the process ends once distillate is drawn into the distillate receiver V003. The withdrawal of distillate begins when the temperatures in the distillation column stop shifting significantly over time.

During the operation phase, the condensate in the buffer vessel V002 is split into reflux and product, collected in the distillate receiver V003. Eventually, anomalies are introduced in the operation phase, as described in more detail below. The operation phase ends either after a specified amount of distillate is obtained (or, equivalently, the filling level of the reboiler vessel drops under a certain value) or simply upon time-out. Then, the shut-down phase begins.

For the shut-down, first all heating elements were ramped down, and the pressure control was switched off. Online sampling continued at least until all samples drawn from the process by the inline-sampling pump P201 during the operation phase were measured by NMR. After all temperatures in the plant had returned to ambient, the ventilation valve HV007 was opened, leading to ambient pressure in the plant. The reboiler vessel V001 was then emptied and rinsed with suitable cleaning agents, such as 2-propanol or purified water. Moreover, the buffer vessel V002 and the distillate receiver V003 were emptied. The NMR sample line was rinsed and emptied. Finally, the plant was sealed and dried under vacuum for 6 to 12 hours using the vacuum pump P301. The plant is considered to be sufficiently clean if it passes the leakage test in the following experiment. The described procedure was applied when the mixture was changed in the following experiment. If the same mixture was studied, a simplified cleaning procedure was applied.

\subsubsection{Anomalies}
\label{subsec:BATCH_Anomalies}

Perturbations were intentionally and systematically induced in the operating plant to generate anomalous process data. The perturbations disrupt normal operation, leading to sensor readings that deviate from expected behavior. Any such observable deviation recorded by the sensors is called an anomaly here. The perturbations were introduced at a specific point in time (typically by changing the setting of some actuators) and then maintained for a specific time span before being removed again (typically by resetting the actuators to their initial values). 
A variety of perturbations were induced, and their timing and duration were varied. They include well-known industrial failure causes~\cite{Kister2003} and can be categorized as follows: (i) Setpoint changes of actuators, e.g., heaters, thermostats, pumps, or automatic valves, (ii) compromising sensor data, e.g., by noise or drift, and (iii) addition or removal of substances from the plant, e.g., foaming agents or nitrogen. An overview of the induced perturbations is given in Table~\ref{table:FailureModes}.

\begin{table}[H]
\caption{Overview of the induced perturbations. For the three categories, the target of the perturbation and the manipulated components are reported.}
\centering
\begin{tabular}{@{}lll@{}}
\hline
Category & Target & Component \\
\hline \hline
\makecell[tl]{Setpoint changes \\of actuators} & \makecell[tl]{Main heat supply to \\reboiler vessel (V001)}
 & H002, H701 \\
\addlinespace[2pt]
& \makecell[tl]{Heat supply to compensate \\heat losses in V001, C001-C003}
  & \makecell[tl]{H702, H706, H704, \\H708} \\
\addlinespace[2pt]
& Condenser cooling (HE001, HE002) & AV716, TCU1 \\
\addlinespace[2pt]
& Pressure control  & TV1 \\
\addlinespace[2pt]
&  Reflux ratio & P701, P702 \\
\addlinespace[4pt]
\makecell[tl]{Compromising \\sensor data} & \makecell[tl]{Control loop between \\sensor and actuator}
  & \makecell[tl]{PDI701, PDI702, T703, \\T705, T709, T711, T712} \\
\addlinespace[4pt]
\makecell[tl]{Addition or removal\\of substances} & \makecell[tl]{System boundary\\ Fluid dynamics}
  & \makecell[tl]{HV001-HV005, HV009\\ V001, C001-C003} \\
\hline
\end{tabular}
\label{table:FailureModes}
\end{table}

For most experiments with an anomaly, we have also carried out a corresponding fault-free experiment. Furthermore, unintended anomalies occurred in a few experiments. These experiments are also included in the dataset, but they are not discussed in the present section.  

\begin{figure}[H]
    \centering
    \includegraphics[width=0.95\textwidth]{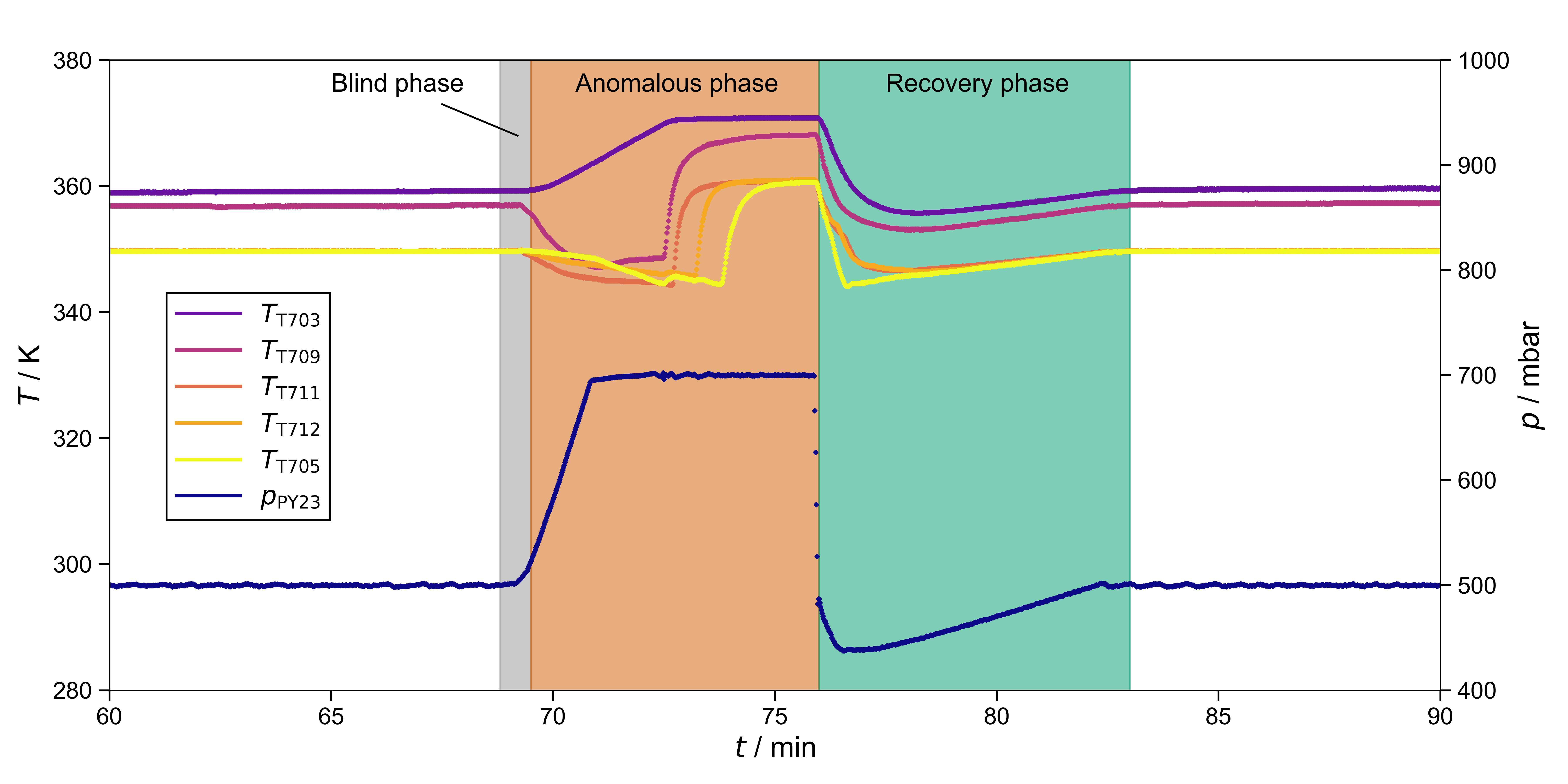}
    \captionsetup{justification=raggedright, singlelinecheck=false}
    \caption{Temperatures and pressure recorded during experiment \nolinkurl{batch_dist_ternary_butan-1-ol+propan-2-ol+water/operating_point_002/test_anormal_experiment_003}, in which a perturbation was introduced at $t=69$ min and removed at $t=76$ min. The colors indicate the three phases of the response to the perturbation. The perturbation was a setpoint change of the pressure from $p=500$ mbar to $p=700$ mbar.}
    \label{fig:Batch_AnomalyScheme}
\end{figure}

\begin{figure}[H]
    \centering
    \includegraphics[width=0.95\linewidth]{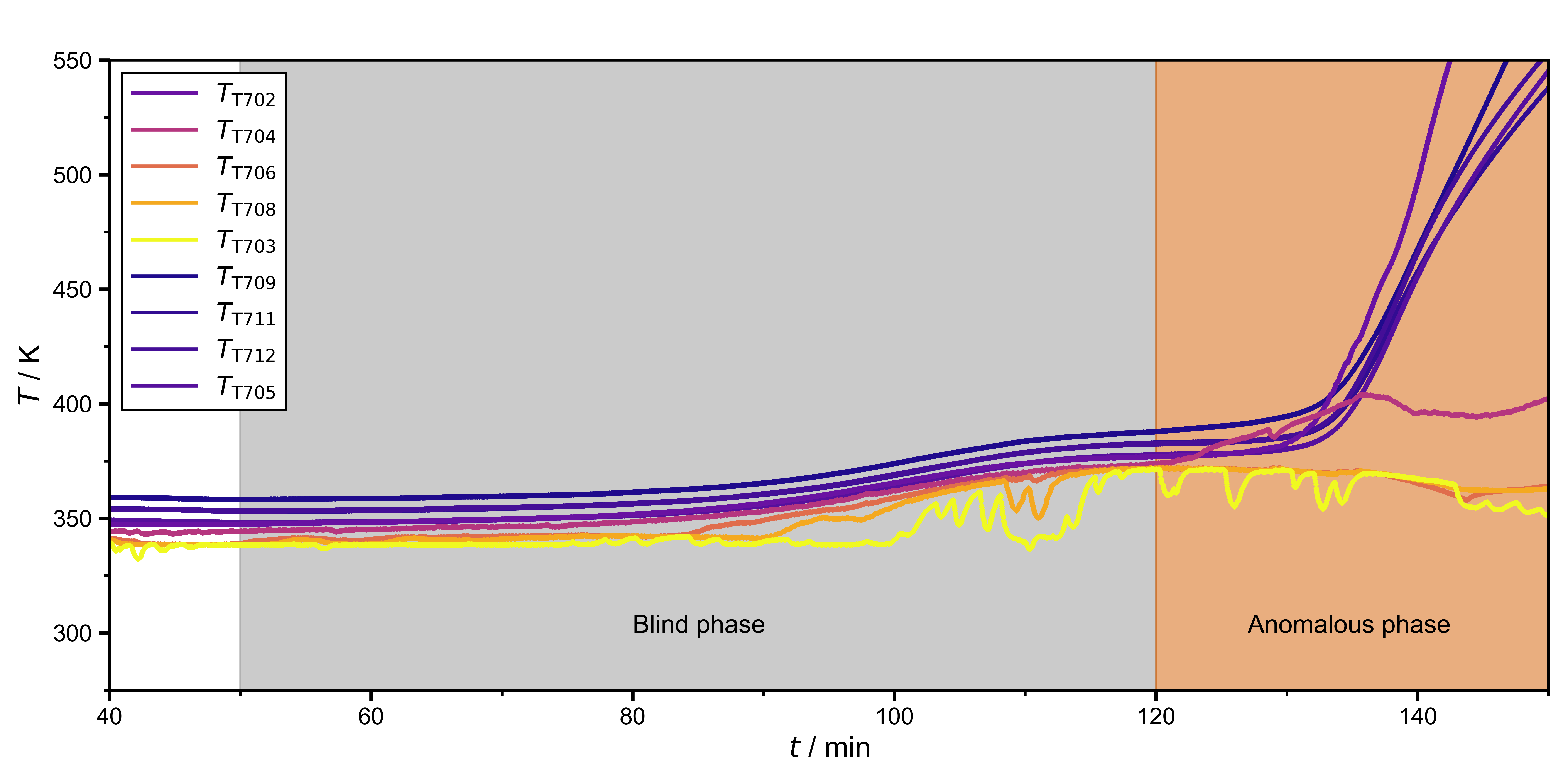}
    \captionsetup{justification=justified, singlelinecheck=false}
    \caption{Temperatures recorded during experiment \nolinkurl{batch_dist_ternary_butan-1-ol+propan-2-ol+water/operating_point_013/test_anormal_experiment_001} in which a perturbation was introduced at $t=50$ min. The process did not recover from the perturbation, which was a fault in the cooling water supply that resulted in the dry-running of the reboiler vessel V001. Hence, the process needed to be shut down. The colors indicate the two occurring phases of the response to the perturbation.}
    \label{fig:BATCH_catastrophic}
\end{figure}

In most cases, the plant recovered after the perturbation was removed. In these cases, the response of the plant to the perturbation can be split into three phases (see Figure~\ref{fig:Batch_AnomalyScheme} for an example): 
\begin{itemize}
    \item Blind phase: The perturbation is induced, but its effects are not yet observable.
    \item Anomalous phase: This phase begins when the anomaly becomes observable in any of the sensor readings and ends when the perturbation is removed.
    \item Recovery phase: This phase begins when the perturbation is removed and ends when its effects are no longer observable.
\end{itemize}
Depending on the experiment, the time the perturbation was applied ranged from 2 to 20 minutes, and the time it took the plant to recover ranged from a few seconds to half an hour. 
Furthermore, the dataset includes a few experiments in which the perturbation remained active until the end of the run. An example is shown in Figure~\ref{fig:BATCH_catastrophic}. 

In most cases, the perturbation were induced deliberately; therefore, the starting time of the blind phase and the end time of the anomalous phase are known. The time at which a perturbation becomes observable in the recorded data, and the time at which its effects are no longer observable were determined by an expert: all time-series data were evaluated after the end of the experiment by one of the authors, who was also experienced in operating the plant, and flagged the points based on their experience and judgment.

\subsection{Metadata of Anomalies}
\label{subsec:BATCH_Metadata_Anomalies}

Contextual metadata is essential for understanding anomalies and ensuring transparency. While raw process signals capture responses to perturbations, metadata explain why these responses occur and how they relate to the underlying plant behavior. In addition to the experimental data, the dataset therefore includes structured metadata describing each labeled anomaly. To provide a standardized, semantically grounded representation, these metadata are formalized using an ontology. Semantic Web Technologies (SWTs)~\cite{ontologies}, with ontologies as a core component, offer a framework for capturing and organizing such knowledge in a uniform, machine-interpretable way. Ontologies define classes and relationships that represent domain knowledge; by instantiating these with concrete details about anomalies and their causes, the metadata become both consistent and interoperable.

The description of perturbations and their effects in this dataset follows the principles of Failure Mode and Effects Analysis (FMEA)~\cite{Burge2010}. Here, the term failure is used interchangeably with perturbation, referring to a disturbance of the process. The adapted FMEA approach distinguishes between the perturbation mode (the manner in which a system is disturbed), the perturbation itself (the initiating event leading to that mode), and the perturbation effect (its immediate consequence)~\cite{Mueller2020,Pecht2009}. The ontology applied in this work builds primarily on the Folio Ontology developed by Steenwinckel et al.~\cite{folio}, extended with concepts introduced by Klein et al.~\cite{Klein2025}. The Folio Ontology aligns the FMEA perspective with the Semantic Sensor Network (SSN)~\cite{ssn} and the Sensor, Observation, Sample, and Actuator (SOSA) ontologies~\cite{sosa}, allowing each perturbation mode to be linked both to the affected system component and to the sensor that records its observable effect. Klein et al.~\cite{Klein2025} further emphasize that a component’s perturbation mode is determined by its function.

The metadata are stored in YAML format (\url{https://yaml.org/}), providing a machine-interpretable yet human-readable representation. Figure~\ref{fig:exampleYAML} shows an example of such a YAML instance, describing the anomaly caused by a temporary pressure increase, cf.~Figure~\ref{fig:Batch_AnomalyScheme}. The example is explained in detail in the following.

\begin{figure}[H]
\begin{tcolorbox}[listing only, colback=white, colframe=black]
\begin{lstlisting}[style=ymlstyle]
 anomalies:
   - class: ConfirmedAnomaly
     id: OP003TA002A001
     hasBeginning: '14:55:31'
     hasEnd: '14:58:00'
     hasCategory: transient
     hasRecoveryAction: Restore normal state
     Perturbation: Increase in pressure setpoint PY23 by 200 mbar to 700 mbar
     PerturbationMode:
       label: Increased vacuum line throttling
       hasBeginning: '14:55:05'
       hasEnd: '14:58:00'
       happenedAt:
         System:
           label: Vacuum Throttle Valve
           id: TV1
           hasFunction: Throttling of vacuum line for system pressure regulation
           hasNormalState:
           - automatic closed-loop controlled
           - pressure setpoint 500 mbar
           - sensor PY23
           hasAnomalousState:
           - automatic closed-loop controlled
           - pressure setpoint 700 mbar
           - sensor PY23
        hasLocalEffect:
            LocalEffect:
              label: Increased system pressure
              id: Observation001_OP002TA003A001
              madeBySensor:
                Sensor: 
                    id: PY23
              observedProperty:
                ObservableProperty: Pressure
              hasFeatureOfInterest:
                FeatureOfInterest: Entire plant
\end{lstlisting}
\end{tcolorbox}
\captionof{figure}{Metadata stored in YAML format summarizing information on an anomaly caused by a temporary pressure increase.}
\label{fig:exampleYAML}
\end{figure}

In the example shown in Figure~\ref{fig:exampleYAML}, the perturbation is a temporary increase of the system pressure setpoint. This perturbation disturbs the system’s normal function and manifests as a specific perturbation mode, here, an increased flow resistance in the vacuum line caused by the throttle valve within the pressure control loop. The throttle valve thus represents the affected system component. Operating in this mode produces a visible anomaly, which is linked to the corresponding sensor observations. Observations follow the SOSA pattern: each observation is made by a sensor, measures an observable property, and pertains to a feature of interest, i.e., the part of the plant whose property is being measured. In this example, the increased throttling of the vacuum line leads to a rise in system pressure, observed by a pressure sensor. Because this change affects the global pressure balance, the feature of interest is not limited to a specific component, such as a pipe segment or vessel, but encompasses the entire plant. A perturbation mode can also give rise to a local effect, representing a system-specific observable response that accompanies its onset. The concept of a local effect distinguishes these targeted, component-level observations from more general observations that may occur over time, also depending on the duration of the perturbation mode. Note that, in some cases, a perturbation may remain unobserved in the available sensor data. On the other hand, anomalies, describing observable deviations from normal plant operation, can be caused by both unexpected sensor deviations without an identifiable cause and be linked to a known perturbation that can be explicitly described (so-called confirmed anomalies); the above described example is a confirmed anomaly. 

\begin{figure}[H]
    \centering
    \includegraphics[width=0.95\linewidth]{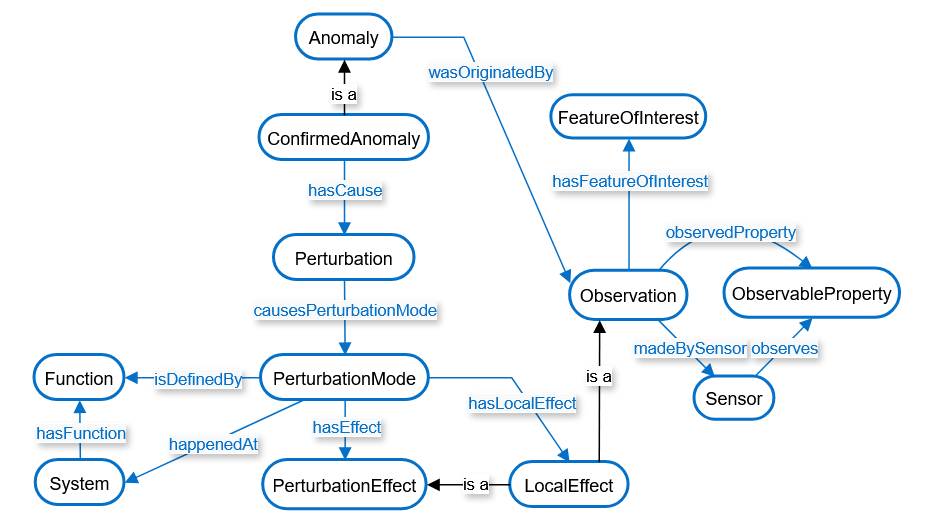}
    \captionsetup{justification=raggedright, singlelinecheck=false}
    \caption{Ontology schema with classes (boxes) and labeled relations (arrows). Blue arrows mark semantic relationships between classes and black arrows denote hierarchical “is a” relations.}
    \label{fig:AnomalySchema}
\end{figure}

The ontology schema underlying the YAML instantiation is shown in Figure~\ref{fig:AnomalySchema}. Blue boxes denote ontology classes, labeled blue arrows indicate semantic relationships, and black arrows denote hierarchical “is a” relations. The directionality of blue arrows in the figure is limited to one side for clarity; however, the actual relations can be bidirectional. For example, the relation “hasCause” between a confirmed anomaly and a perturbation is conceptually complemented by the inverse relation “isCauseOf”.

In addition to semantic relations between the classes, those classes can also have descriptive attributes. A confirmed anomaly includes a recovery action, which describes how the system is restored to its normal condition. It also includes a category that denotes the temporal persistence of an anomaly and is classified as transient, recurrent, or permanent. Transient, thereby, refers to a continuous anomaly that resolves once the perturbation is removed. Recurrent indicates that a single perturbation causes an intermittent anomaly that alternates between observable and non-observable states over time. Permanent characterizes cases where the anomaly persists until the end of the experiment or does not recover even after the perturbation has been resolved. Both anomaly and perturbation mode include beginning and end attributes to record their temporal boundaries. For a perturbation mode during a confirmed anomaly, these timestamps cover the blind phase (called phase “1” in the label modality here) and anomalous phase (called phase “2” in the label modality), whereas for an observable anomaly, they span the anomalous phase and the recovery phase (called phase “3” in the label modality).

The system class includes attributes for its normal and anomalous states, characterizing fault-free and faulty conditions. Every instance of an anomaly, perturbation mode, sensor, observation, and system is identified by a globally unique ID. Sensor and system identifiers correspond to their names in the P\&I diagrams; sensor identifiers also match the names of their data streams. An anomaly’s identifier is unique to that specific event. The attributes are summarized in Table~\ref{tab:class_attributes}.

\begin{table}[H]
    \caption{Class attributes introduced in the dataset.}
  \centering
  \begin{tabular}{@{}ll@{}}
    \hline
    Class  & Attribute \\
    \hline \hline
    Anomaly, Perturbation mode & \emph{hasBeginning}, \emph{hasEnd}, \emph{id} \\
    Confirmed anomaly & \emph{hasCategory} \\
    System & \emph{hasNormalState}, \emph{hasAnomalousState}, \emph{id} \\
    Observation, Sensor & \emph{id} \\
    Confirmed anomaly & \emph{hasRecoveryAction} \\
    \hline
  \end{tabular}
  \label{tab:class_attributes}
\end{table}

Each ontology class is instantiated with plain-text values rather than a fixed hierarchy of predefined phrases. This deliberately lightweight design keeps the representation flexible while remaining semantically grounded. For the dataset presented here, domain experts annotated each anomaly according to the ontology’s structure and through careful visual inspection of the corresponding sensor readings. 

Each YAML file contains the anomaly instances of a single experiment. Anomalies are listed in chronological order, and filenames encode both the operating point and the anomalous experiment.  
As an example, \nolinkurl{OP003TA002A001} identifies the first anomaly (\texttt{001}) associated with the second time series for operating point 3 (file: \nolinkurl{operating_point_003}/\nolinkurl{test_anormal_experiment_002.csv}). Observation IDs are sequential and reference their associated anomaly; for instance, \nolinkurl{Observation001_OP003TA002A001} denotes the first observation linked to anomaly \nolinkurl{OP003TA002A001}. When multiple components were perturbed simultaneously, each component exhibited its own perturbation mode. However, these modes are grouped under a single anomaly ID representing the shared perturbation, e.g., a simultaneous shutdown of all mantle heaters. Currently, only the local effect of each mode is documented, describing its immediate, component-specific impact. This structure can be readily expanded to capture more complex relationships, such as causal chains and the propagation of anomalies through the system.

This explicit mapping renders the metadata both machine-interpretable and human-readable. Each anomaly thus becomes not only a labeled segment of sensor data but also a coherent knowledge unit describing what happened, why it happened, and how it was observed. It therefore provides a valuable foundation for future applications, particularly for developing AD methods that not only identify anomalies but also infer their underlying causes.

\section{Data Record}
\label{sec:BATCH_datamanagement}
\subsection{Overview of Experimental Data and the Metadata}
The complete dataset described in this work is publicly available in a structured format under a CC BY 4.0 license at \rurl{doi.org/10.5281/zenodo.17395543}\cite{MainDataset}. During each experiment, time-series data are collected. Each experiment generates time-series data over a period of 1-8 h, with a sampling rate of 1 s. In the dataset, we provide actuator and sensor data of each experiment in separate files. For each sensor, information on the measurement uncertainties is also provided. The time-series data is labeled to distinguish start-up, operation, and shut-down. Missing data points, e.g., for sensors and actuators due to scarce timing issues in the PCS (relevant for about one percent of the data), were interpolated linearly, resulting in .csv files containing time-series data from 18 sensors and 13 actuators for each second of an experiment. A flag indicating, if a timestep is interpolated or not is provided in the measurement uncertainty modality. The time-series anomaly labels for the operation are provided in separate .csv files. Structured metadata provide context for the anomalous process data by specifying the occurring anomalies as described in Section~\nameref{subsec:BATCH_Metadata_Anomalies}.
Moreover, liquid density, pressure, and mass flux data obtained from the NMR sample pump P201 are available, which were, however, not considered as sensor data as input for the AD methods, but were only used to assign the concentrations measured by online NMR spectroscopy to the correct time step in the time series, cf. Supplementary Information for details. 
In addition to the time-series data described above, several other data modalities were collected with the unconventional sensors described in Section~\nameref{sec:BATCH_Plant}. Operating-phase audio data are provided as a series of .wav files, each 40 seconds long. Images from the cameras described in Section~\nameref{sec:BATCH_Plant} are compressed into .mp4 video files. Concentration data measured with online NMR and offline gas chromatography are supplied as raw tabular and processed time-series data. The supplied information enables assigning the point in time in the experiment to the data in all cases.
Furthermore, for each experiment, tabular data sheets on the ambient conditions, the detailed plant control regime, and the log files from the PCS described in Section~\nameref{subsec:BATCH_Control} are available. Moreover, general substance data for the respective systems, including residue curves and VLE behavior, are provided, along with tabular information on the initial mixture for each experiment.

\subsection{Dataset Structure}
\begin{figure}[H]
    \centering
    \includegraphics[width=0.95\linewidth]{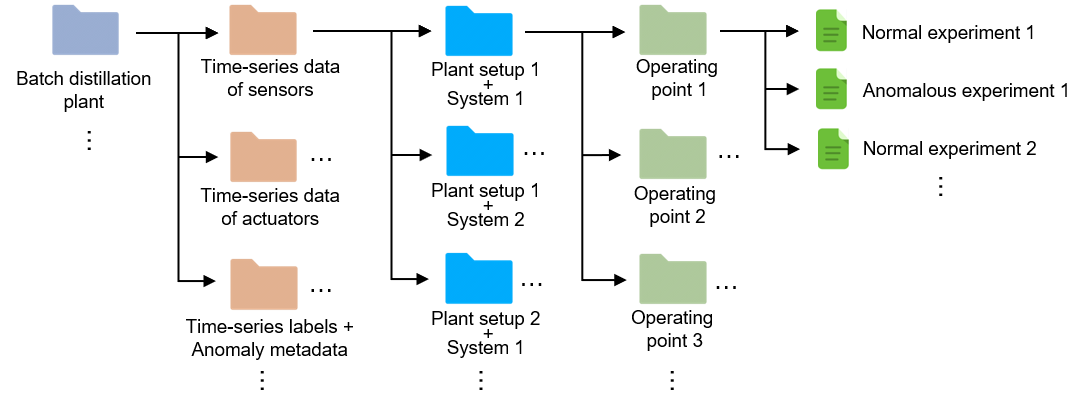}
    \captionsetup{justification=raggedright, singlelinecheck=false}
    \caption{Overview of the structure of the dataset created in the present work. The hierarchy level decreases from left to right. The colored folders indicate the classes on each level. The actual data are located on lowest level.}
    \label{fig:DataStructure}
\end{figure}

In the dataset data of different modalities, i.e., different types of information, are stored. The dataset is structured into several levels, as shown in Figure~\ref{fig:DataStructure}. On the top level, general metadata on the plant is provided, including the information given in Section~\nameref{sec:BATCH_Plant} of this paper. The following levels correspond to subdirectories in the dataset's file structure. The actual data are found on the lowest level. 

On the second level, the data are split according to their modality: 
times-series data of sensors, actuators, labels (+ anomaly metadata), audio, image, NMR, GC, and sensor uncertainties as well as metadata on the plant units, operation logs, ambient conditions, and the substances and mixtures.
The reason for splitting the data from the same experiment by their modalities at this high level is that, in many applications, data from certain modalities (e.g., audio, video) are not of interest. Furthermore, not all modalities are available for every experiment; an overview is given in the Supplementary Information. 

On the third level, different combinations (plant setups + systems) are distinguished. Setups are defined by function and describe the plant’s actual configuration, which varies only slightly -- mainly when sensors or minor components are changed. At the fourth level, the operating point is specified, i.e., the set of all settings used in the plant operation, including parameters, controller setpoints, and fixed device power outputs. At the lowest level, the actual individual datasets are found, including both normal and anomalous runs. Anomalous runs are accompanied by metadata describing the anomaly and its cause, and both normal and anomalous experiments may include additional comments as metadata.

Hence, each individual dataset belongs to a directory tree along which all pertinent information on the experiment can be obtained. The tree structure also enables an unambiguous assignment of individual datasets of different modalities (e.g., sensor and actuator) to an experiment. This assignment is also facilitated by the filenames, which always contain information about the experiment. 

\section{Data Overview}
In total, the dataset includes process data from 119 experiments, comprising more than one million time steps. Table~\ref{tab:BATCH_experiments} gives an overview of the number of experiments and  operating points for each studied system. The number of experiments often exceeds the number of operating points because, in many cases, especially when anomalies were introduced, the same operating point was chosen in different runs. Experiments with the system (ethanol + 2-propanol) were conducted during preliminary plant tests. Mixtures, substance purities, and residue-curve plots are provided in the Supplementary Information and in the dataset as metadata. The time-series data are partitioned into start-up, operation, and shut-down phases; anomaly labels and metadata are available only for the operation phase.

\begin{table}[H]
    \centering
    \caption{Overview of normal and anomalous experiments per system, including the number of experiments and unique operating points; operating points may repeat across experiments.}
    \begin{tabular}{llcc}
        \hline
        System & Type & \makecell[tl]{No. of\\experiments} & \makecell[tl]{No. of\\ operating points} \\ 
        \hline \hline
        1-Butanol + 2-propanol + water & Normal & 29 & 25 \\
                            & Anomalous & 62 & 27\\
        Acetone + 1-butanol + methanol & Normal & 8 & 8 \\
                            & Anomalous &  16  & 12 \\
        Ethanol + 2-propanol & Normal & 3 & 2 \\
                            & Anomalous & 1 & 1 \\
        \hline
    \end{tabular}
    \label{tab:BATCH_experiments}
\end{table}

\paragraph{Examples.}
\label{subsec:BATCH_dataexamples}
In Figure~\ref{fig:Batch_Example_1}, data from an anomalous experiment is compared to data from the corresponding normal experiment. 
The figure shows results for the system (1-butanol + 2-propanol + water) and presents data for the temperatures along the column and the pressure, as well as the composition of the mixture in the reboiler vessel V001 measured by NMR spectroscopy. Further sensor data for these experiments are provided in the Supplementary Information. We start the discussion with the results from the normal experiment (upper two panels in Figure~\ref{fig:Batch_Example_1}). During start-up, the temperature in the reboiler vessel (T703) increases rapidly and shows a kink when the boiling temperature of the feed is reached, as expected. The temperature measured below the first column section (T709) is similar to the temperature in the reboiler vessel (T703) throughout the experiment. The onset of evaporation also leads to a jump in the temperatures measured below the other column sections (T711, T712) and the temperature at the top of the column (T705). During the experiment, these temperatures remain relatively constant first, which is a result of the weak influence of the changes in the composition on the temperature. They begin to rise significantly when the mixture in the reboiler vessel V001 is almost depleted of water. At the end of the run, a small amount of highly purified 1-butanol, the heavy boiler in the studied system, remains in the reboiler vessel V001. The pressure is constant throughout the experiment.

The initial settings of the anomalous experiment (lower two panels in Figure~\ref{fig:Batch_Example_1}) were the same as those for the normal experiment, which is why the signals obtained from both experiments are generally similar. There are two perturbations, which lead to anomalies and which can be discerned easily. The three phases of the perturbation (blind phase, anomalous phase, recovery phase) are marked in each case as described in Section~\nameref{subsec:BATCH_Anomalies}. The perturbations consisted of changes in the setpoint of the heating H002 in the reboiler vessel V001 (normal setpoint: 105~W; first perturbation: temporary increase to 205~W; second perturbation: temporary decrease to 35~W).

\begin{figure}[H]
    \centering
    \includegraphics[width=\textwidth]{Figures_Submission/PAPER_Revised_Batch_Example_1.png}
    \captionsetup{justification=justified, singlelinecheck=false}
    \caption{Two experiments carried out with the batch distillation plant. The upper two panels show results from a normal experiment (\nolinkurl{batch_dist_ternary_butan-1-ol+propan-2-ol+water/operating_point_028/train_normal_experiment_001}), the lower two panels show the results from a corresponding anomalous experiment with two perturbations (\nolinkurl{batch_dist_ternary_butan-1-ol+propan-2-ol+water/operating_point_028/test_anormal_experiment_001}). The colors indicate the three phases of the response to the perturbation. Both perturbations consist of changes of the setpoint of heating H002 in the reboiler vessel V001 (see text for details). In both experiments, batch distillation was performed at 700~mbar, a heat supply of 105~W, and a reflux ratio of 2.}    
    \label{fig:Batch_Example_1}
\end{figure}

\section{Technical Validation}
\label{sec:technical_validation}
To assess the suitability of the dataset for the development of AD methods in chemical process engineering, it was used to train and test multiple time-series AD methods from the contemporary literature~\cite{Wagner2023}. The studied AD methods include forecasting-based methods~\cite{malhotra2015a}, reconstruction-based methods~\cite{kim2022a}, generative methods~\cite{soelch2016a}, and hybrid approaches~\cite{shen2020a, zhao2020multivariate}, which have already been evaluated on the synthetic TEP dataset~\cite{Rieth2017, Downs1993} before~\cite{Hartung2023}. All studied AD methods are available in an open-source git-repository~\cite{Wagner2023}. 

In this work, we have trained the AD methods on sensor data from a subset of fault-free experiments in our dataset, namely, the experiments \nolinkurl{batch_dist_ternary_butan-1-ol+propan-2-ol+water/operating_point_001} to \nolinkurl{..003}. 
For training we used 1000 time steps from each \texttt{normal} experiment of these operating points. As test set, all \texttt{anormal} experiments of these operating points were used. All training data sets were normalized and anomaly-free. The hyperparameters were not fixed, but optimized using five-fold cross-validation. Hyperparameter optimization regarding the $AUPRC$ score (see below) was performed for 24 hours computing time for each method using the \texttt{TimeSeAD} python package~\cite{Wagner2023}. The results for the TEP dataset were created in the same manner.
Using the trained model with the best hyperparameter set, the tests were conducted using the sensor data and anomaly labels provided for the anomalous experiments of this subset. As the performance metric, we calculated the area under the precision-recall curve ($AUPRC$), an established metric for assessing the classification performance, with the precision and recall defined as
\begin{equation}
    Precision(\tau) = \frac{t_\mathrm{p}(\tau)}{t_{\mathrm{p}}(\tau)+f_{\mathrm{p}}(\tau)}
\end{equation}
and
\begin{equation}
    Recall(\tau) = \frac{t_{\mathrm{p}}(\tau)}{t_{\mathrm{p}}(\tau)+f_{\mathrm{n}}(\tau)}
\end{equation}
where $t_{\mathrm{p}}$ denotes true positives, $f_{\mathrm{p}}$ false positives, and $f_\mathrm{n}$ false negatives, which are all functions of the decision threshold $\tau$.

\begin{figure}[H]
    \centering
    \includegraphics[width=0.9\linewidth]{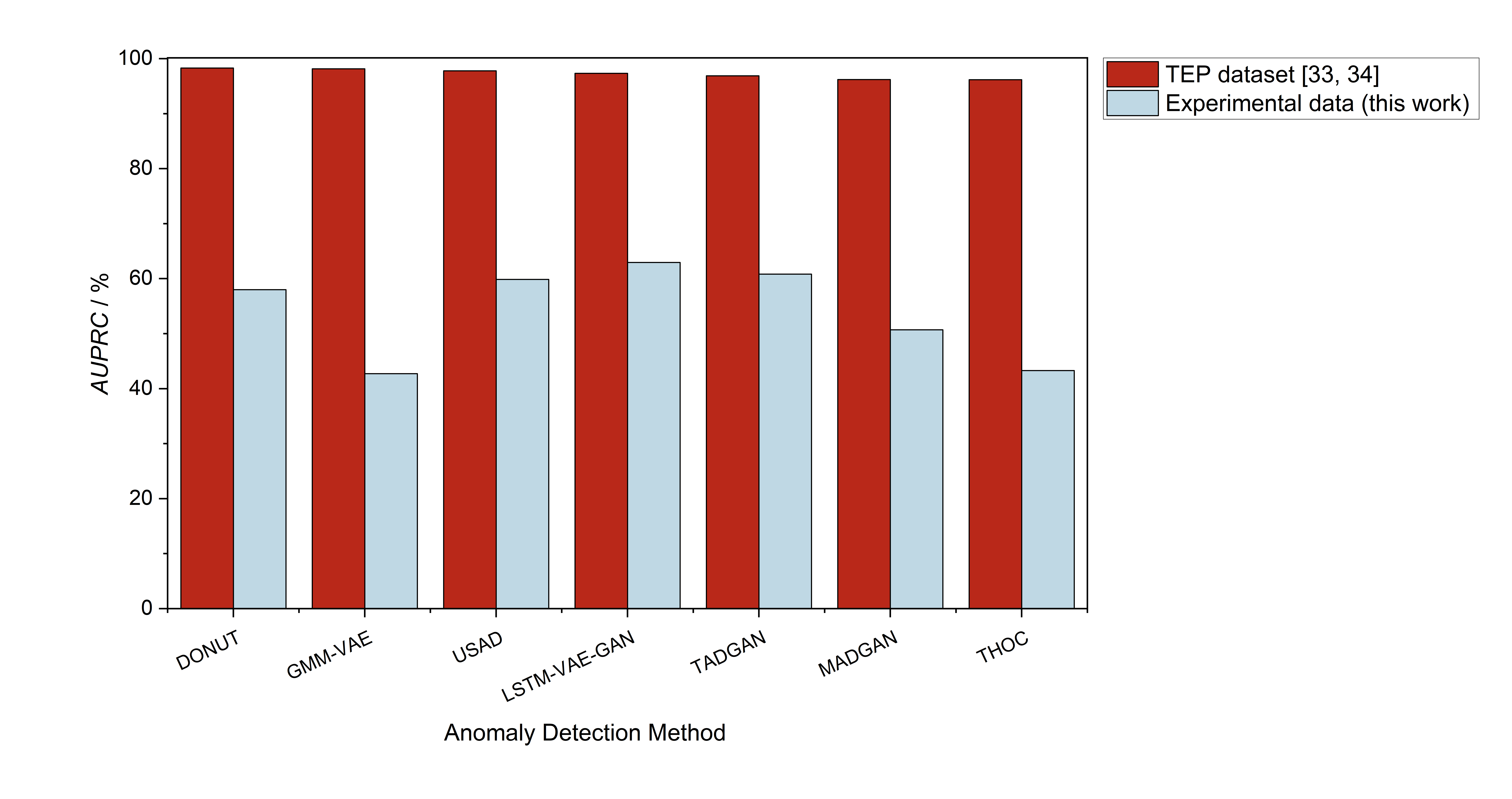}
    \caption{$AUPRC$ scores for multiple AD methods trained and tested on the synthetic TEP dataset~\cite{Rieth2017} and a subset of the experimental dataset presented in this work. The poor performance of all AD methods on the experimental data here provides evidence that the dataset is complex and suitable for the development and testing of AD methods for chemical processes.}
    \label{fig:technical_validation}
\end{figure}

Figure~\ref{fig:technical_validation} shows the performance of the studied AD methods on both the TEP data and the subset of the experimental data from this work using the $AUPRC$. The results show excellent performance on synthetic TEP data but significantly inferior performance on experimental process data. These results demonstrate two key points: first, the experimental batch distillation dataset provided in this work is well-suited for training, testing, and comparing AD methods. Second, the provided experimental dataset is significantly more complex than the established TEP dataset, making it a highly valuable resource for developing and benchmarking advanced AD methods in future work.

\section{Usage Notes}
\label{sec:usage_notes}
This dataset is intended to facilitate research on data-driven monitoring and AD in chemical processes. It is suitable for benchmarking supervised and unsupervised AD methods, developing interpretable and explainable ML models for anomaly diagnosis, and for data-driven strategies for anomaly mitigation and control. The multi-modal nature of the data allows studies on sensor fusion, temporal alignment, and uncertainty-aware learning.

All runs include synchronized multivariate time series, experiment-level metadata, and (when applicable) structured anomaly annotations. Users should rely on the metadata files to construct subsets tailored to specific tasks, such as separating normal from anomalous operation, selecting particular anomaly types, or grouping runs by operating conditions.
Before model development, users should examine sampling intervals, units, and potential missing values. Standard preprocessing steps—such as normalization, handling missing segments, or removing strongly imbalanced variables—may be required depending on the chosen modeling approach. For benchmarking classification-oriented methods, we recommend splitting data by entire experiments rather than by individual time points to avoid temporal leakage and overly optimistic performance estimates.

Anomaly labels indicate the occurrence of anomalies, and the ontology and metadata foster standardized representations of process anomalies. Because the timing of anomaly boundaries is approximate and based on best engineering judgment, users may consider allowing small temporal tolerances when evaluating frame-wise detection methods. For unsupervised anomaly detection, labels should be used strictly for evaluation.

The data can be accessed using standard scientific computing tools. When publishing results based on the dataset, users should cite the dataset DOI and clearly state which subset and version were used.

\section*{Data Availability}
\label{sec:Dataavailability}
The dataset is available in an open Zenodo repository (\rurl{doi.org/10.5281/zenodo.17395543})~\cite{MainDataset}. The Zenodo repository contains the data structured as described in this paper, in .csv, .mp4, and .wav files. Additionally, it contains metadata about the anomalies in a structured .yaml file format.

\section*{Code Availability}
\label{sec:Code_Availability}
The code relevant for this work is available at the public repository: \rurl{github.com/Jarweile/batch-distillation-anomaly-detection-data} (Github). The exact version used in this manuscript is archived at Zenodo (DOI: \rurl{doi.org/10.5281/zenodo.18744032}) corresponding to release \texttt{v1.0.1} (commit: \texttt{bae1683}). 

\section*{Funding}
\label{sec:BATCH_Acknowledgements}
The authors gratefully acknowledge funding from the Deutsche Forschungsgemeinschaft (DFG), as this work was conducted within the DFG Research Unit FOR 5359 "Deep Learning on Sparse Chemical Process Data" (grant number 459419731). Moreover, we thank the DFG Gerätezentrum "Laboratory for Advanced Spin Engineering – Magnetic Resonance (LASE-MR)" (grant number 537627671) for its technical support.

\section*{Supplementary Information}
Supplementary information is available for this work and includes: 
\begin{itemize}
    \item Additional information on the batch distillation plant
    \item Information on sensor calibration
    \item Information on ambient data collection
    \item Information on offline and online concentration measurement
    \item Additional information on chemical compounds and mixtures
\end{itemize}
\newpage

\begin{mcitethebibliography}{55}
\providecommand*\natexlab[1]{#1}
\providecommand*\mciteSetBstSublistMode[1]{}
\providecommand*\mciteSetBstMaxWidthForm[2]{}
\providecommand*\mciteBstWouldAddEndPuncttrue
  {\def\EndOfBibitem{\unskip.}}
\providecommand*\mciteBstWouldAddEndPunctfalse
  {\let\EndOfBibitem\relax}
\providecommand*\mciteSetBstMidEndSepPunct[3]{}
\providecommand*\mciteSetBstSublistLabelBeginEnd[3]{}
\providecommand*\EndOfBibitem{}
\mciteSetBstSublistMode{f}
\mciteSetBstMaxWidthForm{subitem}{(\alph{mcitesubitemcount})}
\mciteSetBstSublistLabelBeginEnd
  {\mcitemaxwidthsubitemform\space}
  {\relax}
  {\relax}

\bibitem[Chandola \latin{et~al.}(2009)Chandola, Banerjee, and Kumar]{Chandola2009}
Chandola,~V.; Banerjee,~A.; Kumar,~V. Anomaly detection: A survey. \emph{ACM Comput. Surv.} \textbf{2009}, \emph{41}, 1--58, DOI: \doi{10.1145/1541880.1541882}\relax
\mciteBstWouldAddEndPuncttrue
\mciteSetBstMidEndSepPunct{\mcitedefaultmidpunct}
{\mcitedefaultendpunct}{\mcitedefaultseppunct}\relax
\EndOfBibitem
\bibitem[Venkatasubramanian \latin{et~al.}(2003)Venkatasubramanian, Rengaswamy, Yin, and Kavuri]{Venkatasubramanian2003}
Venkatasubramanian,~V.; Rengaswamy,~R.; Yin,~K.; Kavuri,~S.~N. A review of process fault detection and diagnosis Part I: Quantitative model-based methods. \emph{Comput. Chem. Eng.} \textbf{2003}, \emph{27}, 293--311, DOI: \doi{10.1016/s0098-1354(02)00160-6}\relax
\mciteBstWouldAddEndPuncttrue
\mciteSetBstMidEndSepPunct{\mcitedefaultmidpunct}
{\mcitedefaultendpunct}{\mcitedefaultseppunct}\relax
\EndOfBibitem
\bibitem[Venkatasubramanian \latin{et~al.}(2003)Venkatasubramanian, Rengaswamy, and Kavuri]{Venkatasubramanian2003a}
Venkatasubramanian,~V.; Rengaswamy,~R.; Kavuri,~S.~N. A review of process fault detection and diagnosis Part II: Qualitative models and search strategies. \emph{Comput. Chem. Eng.} \textbf{2003}, \emph{27}, 313--326, DOI: \doi{10.1016/s0098-1354(02)00161-8}\relax
\mciteBstWouldAddEndPuncttrue
\mciteSetBstMidEndSepPunct{\mcitedefaultmidpunct}
{\mcitedefaultendpunct}{\mcitedefaultseppunct}\relax
\EndOfBibitem
\bibitem[Venkatasubramanian \latin{et~al.}(2003)Venkatasubramanian, Rengaswamy, Kavuri, and Yin]{Venkatasubramanian2003b}
Venkatasubramanian,~V.; Rengaswamy,~R.; Kavuri,~S.~N.; Yin,~K. A review of process fault detection and diagnosis Part III: Process history based methods. \emph{Comput. Chem. Eng.} \textbf{2003}, \emph{27}, 327--346, DOI: \doi{10.1016/s0098-1354(02)00162-x}\relax
\mciteBstWouldAddEndPuncttrue
\mciteSetBstMidEndSepPunct{\mcitedefaultmidpunct}
{\mcitedefaultendpunct}{\mcitedefaultseppunct}\relax
\EndOfBibitem
\bibitem[Chiang(2001)]{Chiang2001}
Chiang,~L.~H. In \emph{Fault Detection and Diagnosis in Industrial Systems}; Russell,~E.~L., Braatz,~R.~D., Eds.; Advanced Textbooks in Control and Signal Processing; Springer: London, 2001\relax
\mciteBstWouldAddEndPuncttrue
\mciteSetBstMidEndSepPunct{\mcitedefaultmidpunct}
{\mcitedefaultendpunct}{\mcitedefaultseppunct}\relax
\EndOfBibitem
\bibitem[Hodge and Austin(2004)Hodge, and Austin]{Hodge2004}
Hodge,~V.; Austin,~J. A Survey of Outlier Detection Methodologies. \emph{Artif. Intell. Rev.} \textbf{2004}, \emph{22}, 85--126, DOI: \doi{10.1023/b:aire.0000045502.10941.a9}\relax
\mciteBstWouldAddEndPuncttrue
\mciteSetBstMidEndSepPunct{\mcitedefaultmidpunct}
{\mcitedefaultendpunct}{\mcitedefaultseppunct}\relax
\EndOfBibitem
\bibitem[Mowbray \latin{et~al.}(2022)Mowbray, Vallerio, Perez-Galvan, Zhang, Del Rio~Chanona, and Navarro-Brull]{Mowbray2022}
Mowbray,~M.; Vallerio,~M.; Perez-Galvan,~C.; Zhang,~D.; Del Rio~Chanona,~A.; Navarro-Brull,~F.~J. Industrial data science – a review of machine learning applications for chemical and process industries. \emph{React. Chem. Eng.} \textbf{2022}, \emph{7}, 1471--1509, DOI: \doi{10.1039/d1re00541c}\relax
\mciteBstWouldAddEndPuncttrue
\mciteSetBstMidEndSepPunct{\mcitedefaultmidpunct}
{\mcitedefaultendpunct}{\mcitedefaultseppunct}\relax
\EndOfBibitem
\bibitem[Dobbelaere \latin{et~al.}(2021)Dobbelaere, Plehiers, Van~de Vijver, Stevens, and Van~Geem]{Dobbelaere2021}
Dobbelaere,~M.~R.; Plehiers,~P.~P.; Van~de Vijver,~R.; Stevens,~C.~V.; Van~Geem,~K.~M. Machine Learning in Chemical Engineering: Strengths, Weaknesses, Opportunities, and Threats. \emph{Engineering} \textbf{2021}, \emph{7}, 1201--1211, DOI: \doi{10.1016/j.eng.2021.03.019}\relax
\mciteBstWouldAddEndPuncttrue
\mciteSetBstMidEndSepPunct{\mcitedefaultmidpunct}
{\mcitedefaultendpunct}{\mcitedefaultseppunct}\relax
\EndOfBibitem
\bibitem[Göttl \latin{et~al.}(2025)Göttl, Pirnay, Burger, and Grimm]{Goettl2025}
Göttl,~Q.; Pirnay,~J.; Burger,~J.; Grimm,~D.~G. Deep reinforcement learning enables conceptual design of processes for separating azeotropic mixtures without prior knowledge. \emph{Comput. Chem. Eng.} \textbf{2025}, \emph{194}, 108975, DOI: \doi{10.1016/j.compchemeng.2024.108975}\relax
\mciteBstWouldAddEndPuncttrue
\mciteSetBstMidEndSepPunct{\mcitedefaultmidpunct}
{\mcitedefaultendpunct}{\mcitedefaultseppunct}\relax
\EndOfBibitem
\bibitem[Gond \latin{et~al.}(2025)Gond, Sohns, Leitte, Hasse, and Jirasek]{Gond2025}
Gond,~D.; Sohns,~J.-T.; Leitte,~H.; Hasse,~H.; Jirasek,~F. Hierarchical matrix completion for the prediction of properties of binary mixtures. \emph{Comput. Chem. Eng.} \textbf{2025}, \emph{199}, 109122, DOI: \doi{10.1016/j.compchemeng.2025.109122}\relax
\mciteBstWouldAddEndPuncttrue
\mciteSetBstMidEndSepPunct{\mcitedefaultmidpunct}
{\mcitedefaultendpunct}{\mcitedefaultseppunct}\relax
\EndOfBibitem
\bibitem[Hayer \latin{et~al.}(2022)Hayer, Jirasek, and Hasse]{Hayer2022}
Hayer,~N.; Jirasek,~F.; Hasse,~H. Prediction of Henry’s law constants by matrix completion. \emph{AIChE J.} \textbf{2022}, \emph{68}, DOI: \doi{10.1002/aic.17753}\relax
\mciteBstWouldAddEndPuncttrue
\mciteSetBstMidEndSepPunct{\mcitedefaultmidpunct}
{\mcitedefaultendpunct}{\mcitedefaultseppunct}\relax
\EndOfBibitem
\bibitem[Hayer \latin{et~al.}(2025)Hayer, Hasse, and Jirasek]{Hayer2025}
Hayer,~N.; Hasse,~H.; Jirasek,~F. Modified UNIFAC 2.0-A Group-Contribution Method Completed with Machine Learning. \emph{Ind. Eng. Chem. Res.} \textbf{2025}, \emph{64}, 10304--10313, DOI: \doi{10.1021/acs.iecr.5c00077}\relax
\mciteBstWouldAddEndPuncttrue
\mciteSetBstMidEndSepPunct{\mcitedefaultmidpunct}
{\mcitedefaultendpunct}{\mcitedefaultseppunct}\relax
\EndOfBibitem
\bibitem[Hayer \latin{et~al.}(2025)Hayer, Specht, Arweiler, Hasse, and Jirasek]{Hayer2025a}
Hayer,~N.; Specht,~T.; Arweiler,~J.; Hasse,~H.; Jirasek,~F. Similarity-Informed Matrix Completion Method for Predicting Activity Coefficients. \emph{J. Phys. Chem.} \textbf{2025}, \emph{129}, 3141--3147, DOI: \doi{10.1021/acs.jpca.4c08360}\relax
\mciteBstWouldAddEndPuncttrue
\mciteSetBstMidEndSepPunct{\mcitedefaultmidpunct}
{\mcitedefaultendpunct}{\mcitedefaultseppunct}\relax
\EndOfBibitem
\bibitem[Hayer \latin{et~al.}(2025)Hayer, Wendel, Mandt, Hasse, and Jirasek]{Hayer2025b}
Hayer,~N.; Wendel,~T.; Mandt,~S.; Hasse,~H.; Jirasek,~F. Advancing thermodynamic group-contribution methods by machine learning: UNIFAC 2.0. \emph{Chem. Eng. J.} \textbf{2025}, \emph{504}, 158667, DOI: \doi{10.1016/j.cej.2024.158667}\relax
\mciteBstWouldAddEndPuncttrue
\mciteSetBstMidEndSepPunct{\mcitedefaultmidpunct}
{\mcitedefaultendpunct}{\mcitedefaultseppunct}\relax
\EndOfBibitem
\bibitem[Hoffmann \latin{et~al.}(2025)Hoffmann, Hasse, Jirasek, Hoffmann, Hasse, and Jirasek]{Hoffmann2025}
Hoffmann,~M.; Hasse,~H.; Jirasek,~F.; Hoffmann,~M.; Hasse,~H.; Jirasek,~F. GRAPPA—A hybrid graph neural network for predicting pure component vapor pressures. \emph{Chem. Eng. J. Adv.} \textbf{2025}, \emph{22}, 100750, DOI: \doi{https://doi.org/10.1016/j.ceja.2025.100750}\relax
\mciteBstWouldAddEndPuncttrue
\mciteSetBstMidEndSepPunct{\mcitedefaultmidpunct}
{\mcitedefaultendpunct}{\mcitedefaultseppunct}\relax
\EndOfBibitem
\bibitem[Jirasek and Hasse(2023)Jirasek, and Hasse]{Jirasek2023}
Jirasek,~F.; Hasse,~H. Combining Machine Learning with Physical Knowledge in Thermodynamic Modeling of Fluid Mixtures. \emph{Annu. Rev. Chem. Biomol. Eng.} \textbf{2023}, \emph{14}, 31--51, DOI: \doi{10.1146/annurev-chembioeng-092220-025342}\relax
\mciteBstWouldAddEndPuncttrue
\mciteSetBstMidEndSepPunct{\mcitedefaultmidpunct}
{\mcitedefaultendpunct}{\mcitedefaultseppunct}\relax
\EndOfBibitem
\bibitem[Jirasek \latin{et~al.}(2020)Jirasek, Alves, Damay, Vandermeulen, Bamler, Bortz, Mandt, Kloft, and Hasse]{Jirasek2020}
Jirasek,~F.; Alves,~R. A.~S.; Damay,~J.; Vandermeulen,~R.~A.; Bamler,~R.; Bortz,~M.; Mandt,~S.; Kloft,~M.; Hasse,~H. Machine Learning in Thermodynamics: Prediction of Activity Coefficients by Matrix Completion. \emph{J. Phys. Chem. Lett.} \textbf{2020}, \emph{11}, 981--985, DOI: \doi{10.1021/acs.jpclett.9b03657}\relax
\mciteBstWouldAddEndPuncttrue
\mciteSetBstMidEndSepPunct{\mcitedefaultmidpunct}
{\mcitedefaultendpunct}{\mcitedefaultseppunct}\relax
\EndOfBibitem
\bibitem[Jirasek and Hasse(2021)Jirasek, and Hasse]{Jirasek2021}
Jirasek,~F.; Hasse,~H. Perspective: Machine Learning of Thermophysical Properties. \emph{Fluid Phase Equilib.} \textbf{2021}, \emph{549}, 113206, DOI: \doi{10.1016/j.fluid.2021.113206}\relax
\mciteBstWouldAddEndPuncttrue
\mciteSetBstMidEndSepPunct{\mcitedefaultmidpunct}
{\mcitedefaultendpunct}{\mcitedefaultseppunct}\relax
\EndOfBibitem
\bibitem[Specht \latin{et~al.}(2023)Specht, Arweiler, Stüber, Münnemann, Hasse, and Jirasek]{Specht2023}
Specht,~T.; Arweiler,~J.; Stüber,~J.; Münnemann,~K.; Hasse,~H.; Jirasek,~F. Automated nuclear magnetic resonance fingerprinting of mixtures. \emph{Magn. Reson. Chem.} \textbf{2023}, \emph{62}, 286--297, DOI: \doi{10.1002/mrc.5381}\relax
\mciteBstWouldAddEndPuncttrue
\mciteSetBstMidEndSepPunct{\mcitedefaultmidpunct}
{\mcitedefaultendpunct}{\mcitedefaultseppunct}\relax
\EndOfBibitem
\bibitem[Specht \latin{et~al.}(2024)Specht, Nagda, Fellenz, Mandt, Hasse, and Jirasek]{Specht2024}
Specht,~T.; Nagda,~M.; Fellenz,~S.; Mandt,~S.; Hasse,~H.; Jirasek,~F. HANNA: hard-constraint neural network for consistent activity coefficient prediction. \emph{Chem. Sci.} \textbf{2024}, \emph{15}, 19777--19786, DOI: \doi{10.1039/d4sc05115g}\relax
\mciteBstWouldAddEndPuncttrue
\mciteSetBstMidEndSepPunct{\mcitedefaultmidpunct}
{\mcitedefaultendpunct}{\mcitedefaultseppunct}\relax
\EndOfBibitem
\bibitem[Vollmer \latin{et~al.}(2024)Vollmer, Fellenz, Jirasek, Leitte, and Hasse]{Vollmer2024}
Vollmer,~L.; Fellenz,~S.; Jirasek,~F.; Leitte,~H.; Hasse,~H. KnowTD--An Actionable Knowledge Representation System for Thermodynamics. \emph{J. Chem. Inf. Model.} \textbf{2024}, \emph{64}, 5878--5887, DOI: \doi{10.1021/acs.jcim.4c00647}\relax
\mciteBstWouldAddEndPuncttrue
\mciteSetBstMidEndSepPunct{\mcitedefaultmidpunct}
{\mcitedefaultendpunct}{\mcitedefaultseppunct}\relax
\EndOfBibitem
\bibitem[Chadha \latin{et~al.}(2019)Chadha, Rabbani, and Schwung]{Chadha2019}
Chadha,~G.~S.; Rabbani,~A.; Schwung,~A. Comparison of Semi-supervised Deep Neural Networks for Anomaly Detection in Industrial Processes. Proc. IEEE Int. Conf. Ind. Inform. (INDIN). 2019; pp 214--219, DOI: \doi{10.1109/indin41052.2019.8972172}\relax
\mciteBstWouldAddEndPuncttrue
\mciteSetBstMidEndSepPunct{\mcitedefaultmidpunct}
{\mcitedefaultendpunct}{\mcitedefaultseppunct}\relax
\EndOfBibitem
\bibitem[Monroy \latin{et~al.}(2009)Monroy, Escudero, and Graells]{Monroy2009}
Monroy,~I.; Escudero,~G.; Graells,~M. Anomaly detection in batch chemical processes. ESCAPE-19. 2009; pp 255--260, DOI: \doi{10.1016/s1570-7946(09)70043-4}\relax
\mciteBstWouldAddEndPuncttrue
\mciteSetBstMidEndSepPunct{\mcitedefaultmidpunct}
{\mcitedefaultendpunct}{\mcitedefaultseppunct}\relax
\EndOfBibitem
\bibitem[Inoue \latin{et~al.}(2017)Inoue, Yamagata, Chen, Poskitt, and Sun]{Inoue2017}
Inoue,~J.; Yamagata,~Y.; Chen,~Y.; Poskitt,~C.~M.; Sun,~J. Anomaly Detection for a Water Treatment System Using Unsupervised Machine Learning. Proc. IEEE Int. Conf. Data Min. Workshops (ICDMW). 2017; pp 1058--1065, DOI: \doi{10.1109/icdmw.2017.149}\relax
\mciteBstWouldAddEndPuncttrue
\mciteSetBstMidEndSepPunct{\mcitedefaultmidpunct}
{\mcitedefaultendpunct}{\mcitedefaultseppunct}\relax
\EndOfBibitem
\bibitem[Song and Suh(2019)Song, and Suh]{Song2019}
Song,~B.; Suh,~Y. Narrative texts-based anomaly detection using accident report documents: The case of chemical process safety. \emph{J. Loss Prev. Process Ind.} \textbf{2019}, \emph{57}, 47--54, DOI: \doi{10.1016/j.jlp.2018.08.010}\relax
\mciteBstWouldAddEndPuncttrue
\mciteSetBstMidEndSepPunct{\mcitedefaultmidpunct}
{\mcitedefaultendpunct}{\mcitedefaultseppunct}\relax
\EndOfBibitem
\bibitem[Tian \latin{et~al.}(2020)Tian, Liu, Li, Zhang, and Li]{Tian2020}
Tian,~W.; Liu,~Z.; Li,~L.; Zhang,~S.; Li,~C. Identification of abnormal conditions in high-dimensional chemical process based on feature selection and deep learning. \emph{Chin. J. Chem. Eng.} \textbf{2020}, \emph{28}, 1875--1883, DOI: \doi{10.1016/j.cjche.2020.05.003}\relax
\mciteBstWouldAddEndPuncttrue
\mciteSetBstMidEndSepPunct{\mcitedefaultmidpunct}
{\mcitedefaultendpunct}{\mcitedefaultseppunct}\relax
\EndOfBibitem
\bibitem[Wu \latin{et~al.}(2024)Wu, Zhang, Deng, Zhang, and Chai]{Wu2024}
Wu,~G.; Zhang,~Y.; Deng,~L.; Zhang,~J.; Chai,~T. Cross-Modal Learning for Anomaly Detection in Complex Industrial Process: Methodology and Benchmark. \emph{arXiv} \textbf{2024}, DOI: \doi{10.48550/ARXIV.2406.09016}\relax
\mciteBstWouldAddEndPuncttrue
\mciteSetBstMidEndSepPunct{\mcitedefaultmidpunct}
{\mcitedefaultendpunct}{\mcitedefaultseppunct}\relax
\EndOfBibitem
\bibitem[Hartung \latin{et~al.}(2023)Hartung, Franks, Michels, Wagner, Liznerski, Reithermann, Fellenz, Jirasek, Rudolph, Neider, Leitte, Song, Kloepper, Mandt, Bortz, Burger, Hasse, and Kloft]{Hartung2023}
Hartung,~F. \latin{et~al.}  Deep Anomaly Detection on Tennessee Eastman Process Data. \emph{Chem. Ing. Tech.} \textbf{2023}, \emph{95}, 1077--1082, DOI: \doi{10.1002/cite.202200238}\relax
\mciteBstWouldAddEndPuncttrue
\mciteSetBstMidEndSepPunct{\mcitedefaultmidpunct}
{\mcitedefaultendpunct}{\mcitedefaultseppunct}\relax
\EndOfBibitem
\bibitem[Russell \latin{et~al.}(2000)Russell, Chiang, and Braatz]{Russell2000}
Russell,~E.~L.; Chiang,~L.~H.; Braatz,~R.~D. \emph{Data-driven Methods for Fault Detection and Diagnosis in Chemical Processes}; Springer London, 2000; DOI: \doi{10.1007/978-1-4471-0409-4}\relax
\mciteBstWouldAddEndPuncttrue
\mciteSetBstMidEndSepPunct{\mcitedefaultmidpunct}
{\mcitedefaultendpunct}{\mcitedefaultseppunct}\relax
\EndOfBibitem
\bibitem[Schmidl \latin{et~al.}(2022)Schmidl, Wenig, and Papenbrock]{Schmidl2022}
Schmidl,~S.; Wenig,~P.; Papenbrock,~T. Anomaly detection in time series: a comprehensive evaluation. \emph{Proc. VLDB Endow.} \textbf{2022}, \emph{15}, 1779--1797, DOI: \doi{10.14778/3538598.3538602}\relax
\mciteBstWouldAddEndPuncttrue
\mciteSetBstMidEndSepPunct{\mcitedefaultmidpunct}
{\mcitedefaultendpunct}{\mcitedefaultseppunct}\relax
\EndOfBibitem
\bibitem[Zamanzadeh~Darban \latin{et~al.}(2024)Zamanzadeh~Darban, Webb, Pan, Aggarwal, and Salehi]{Darban2024}
Zamanzadeh~Darban,~Z.; Webb,~G.~I.; Pan,~S.; Aggarwal,~C.; Salehi,~M. Deep Learning for Time Series Anomaly Detection: A Survey. \emph{ACM Comput. Surv.} \textbf{2024}, \emph{57}, 1--42, DOI: \doi{10.1145/3691338}\relax
\mciteBstWouldAddEndPuncttrue
\mciteSetBstMidEndSepPunct{\mcitedefaultmidpunct}
{\mcitedefaultendpunct}{\mcitedefaultseppunct}\relax
\EndOfBibitem
\bibitem[Wang \latin{et~al.}(2025)Wang, Jiang, Zhang, Wei, Xie, and Pang]{Wang2025}
Wang,~F.; Jiang,~Y.; Zhang,~R.; Wei,~A.; Xie,~J.; Pang,~X. A Survey of Deep Anomaly Detection in Multivariate Time Series: Taxonomy, Applications, and Directions. \emph{Sensors} \textbf{2025}, \emph{25}, 190, DOI: \doi{10.3390/s25010190}\relax
\mciteBstWouldAddEndPuncttrue
\mciteSetBstMidEndSepPunct{\mcitedefaultmidpunct}
{\mcitedefaultendpunct}{\mcitedefaultseppunct}\relax
\EndOfBibitem
\bibitem[Downs and Vogel(1993)Downs, and Vogel]{Downs1993}
Downs,~J.; Vogel,~E. A plant-wide industrial process control problem. \emph{Comput. Chem. Eng.} \textbf{1993}, \emph{17}, 245--255, DOI: \doi{10.1016/0098-1354(93)80018-i}\relax
\mciteBstWouldAddEndPuncttrue
\mciteSetBstMidEndSepPunct{\mcitedefaultmidpunct}
{\mcitedefaultendpunct}{\mcitedefaultseppunct}\relax
\EndOfBibitem
\bibitem[Rieth \latin{et~al.}(2017)Rieth, Amsel, Tran, and Cook]{Rieth2017}
Rieth,~C.~A.; Amsel,~B.~D.; Tran,~R.; Cook,~M.~B. Additional Tennessee Eastman Process Simulation Data for Anomaly Detection Evaluation. \emph{Harvard Dataverse} \textbf{2017}, DOI: \doi{10.7910/DVN/6C3JR1}\relax
\mciteBstWouldAddEndPuncttrue
\mciteSetBstMidEndSepPunct{\mcitedefaultmidpunct}
{\mcitedefaultendpunct}{\mcitedefaultseppunct}\relax
\EndOfBibitem
\bibitem[Muraleedharan \latin{et~al.}(2025)Muraleedharan, Ferre, Arweiler, JungJohann, Jirasek, Hasse, and Burger]{APARNA}
Muraleedharan,~A.; Ferre,~A.; Arweiler,~J.; JungJohann,~I.; Jirasek,~F.; Hasse,~H.; Burger,~J. Experimental time series data with and without anomalies from a continuous distillation mini-plant for development of machine learning anomaly detection methods. \emph{engrXiv} \textbf{2025}, DOI: \doi{10.31224/5631}\relax
\mciteBstWouldAddEndPuncttrue
\mciteSetBstMidEndSepPunct{\mcitedefaultmidpunct}
{\mcitedefaultendpunct}{\mcitedefaultseppunct}\relax
\EndOfBibitem
\bibitem[Muraleedharan \latin{et~al.}()Muraleedharan, Ferre, Arweiler, Jungjohann, Jirasek, Hasse, and Burger]{AparnaDataset}
Muraleedharan,~A.; Ferre,~A.; Arweiler,~J.; Jungjohann,~I.; Jirasek,~F.; Hasse,~H.; Burger,~J. Steady state time series data from a continuous distillation plant with and without anomalies for developing machine learning anomaly detection methods. \url{https://zenodo.org/records/17628963}, Version v1\relax
\mciteBstWouldAddEndPuncttrue
\mciteSetBstMidEndSepPunct{\mcitedefaultmidpunct}
{\mcitedefaultendpunct}{\mcitedefaultseppunct}\relax
\EndOfBibitem
\bibitem[Kister(2003)]{Kister2003}
Kister,~H. What Caused Tower Malfunctions in the Last 50 Years? \emph{Chem. Eng. Res. Des.} \textbf{2003}, \emph{81}, 5--26, DOI: \doi{10.1205/026387603321158159}\relax
\mciteBstWouldAddEndPuncttrue
\mciteSetBstMidEndSepPunct{\mcitedefaultmidpunct}
{\mcitedefaultendpunct}{\mcitedefaultseppunct}\relax
\EndOfBibitem
\bibitem[Arweiler \latin{et~al.}()Arweiler, Jungjohann, Muraleedharan, Leitte, Burger, Münnemann, Jirasek, and Hasse]{MainDataset}
Arweiler,~J.; Jungjohann,~I.; Muraleedharan,~A.; Leitte,~H.; Burger,~J.; Münnemann,~K.; Jirasek,~F.; Hasse,~H. Batch Distillation Data for Developing Machine Learning Anomaly Detection Methods. \url{https://zenodo.org/records/17395543}, Version 1.0.1\relax
\mciteBstWouldAddEndPuncttrue
\mciteSetBstMidEndSepPunct{\mcitedefaultmidpunct}
{\mcitedefaultendpunct}{\mcitedefaultseppunct}\relax
\EndOfBibitem
\bibitem[Janowicz \latin{et~al.}(2019)Janowicz, Haller, Cox, {Le Phuoc}, and Lefrançois]{sosa}
Janowicz,~K.; Haller,~A.; Cox,~S.~J.; {Le Phuoc},~D.; Lefrançois,~M. SOSA: A lightweight ontology for sensors, observations, samples, and actuators. \emph{J. Web Semant.} \textbf{2019}, \emph{56}, 1--10, DOI: \doi{https://doi.org/10.1016/j.websem.2018.06.003}\relax
\mciteBstWouldAddEndPuncttrue
\mciteSetBstMidEndSepPunct{\mcitedefaultmidpunct}
{\mcitedefaultendpunct}{\mcitedefaultseppunct}\relax
\EndOfBibitem
\bibitem[Gronle \latin{et~al.}(2014)Gronle, Lyda, Wilke, Kohler, and Osten]{Gronle2014}
Gronle,~M.; Lyda,~W.; Wilke,~M.; Kohler,~C.; Osten,~W. itom: an open source metrology, automation, and data evaluation software. \emph{Appl. Opt.} \textbf{2014}, \emph{53}, 2974, DOI: \doi{10.1364/ao.53.002974}\relax
\mciteBstWouldAddEndPuncttrue
\mciteSetBstMidEndSepPunct{\mcitedefaultmidpunct}
{\mcitedefaultendpunct}{\mcitedefaultseppunct}\relax
\EndOfBibitem
\bibitem[CARPi \latin{et~al.}(2017)CARPi, Minges, and Piel]{Carpi2017}
CARPi,~N.; Minges,~A.; Piel,~M. eLabFTW: An open source laboratory notebook for research labs. \emph{J. Open Source Softw.} \textbf{2017}, \emph{2}, 146, DOI: \doi{10.21105/joss.00146}\relax
\mciteBstWouldAddEndPuncttrue
\mciteSetBstMidEndSepPunct{\mcitedefaultmidpunct}
{\mcitedefaultendpunct}{\mcitedefaultseppunct}\relax
\EndOfBibitem
\bibitem[Ye \latin{et~al.}(2015)Ye, Dasiopoulou, Stevenson, Meditskos, Kontopoulos, Kompatsiaris, and Dobson]{ontologies}
Ye,~J.; Dasiopoulou,~S.; Stevenson,~G.; Meditskos,~G.; Kontopoulos,~E.; Kompatsiaris,~I.; Dobson,~S. Semantic web technologies in pervasive computing: A survey and research roadmap. \emph{Pervasive Mob. Comput.} \textbf{2015}, \emph{23}, 1--25, DOI: \doi{https://doi.org/10.1016/j.pmcj.2014.12.009}\relax
\mciteBstWouldAddEndPuncttrue
\mciteSetBstMidEndSepPunct{\mcitedefaultmidpunct}
{\mcitedefaultendpunct}{\mcitedefaultseppunct}\relax
\EndOfBibitem
\bibitem[Burge(2010)]{Burge2010}
Burge,~S. The Systems Engineering Tool Box. 2010; \url{https://www.burgehugheswalsh.co.uk/Uploaded/1/Documents/FFMEA-Tool-v2.pdf}\relax
\mciteBstWouldAddEndPuncttrue
\mciteSetBstMidEndSepPunct{\mcitedefaultmidpunct}
{\mcitedefaultendpunct}{\mcitedefaultseppunct}\relax
\EndOfBibitem
\bibitem[Müller \latin{et~al.}(2020)Müller, Lunde, and Hönig]{Mueller2020}
Müller,~C.; Lunde,~R.; Hönig,~P. Generation of a Failure Mode and Effects Analysis with SmartIflow. Proc. Eur. Saf. Reliab. Conf.; Proc. Probabilistic Saf. Assess. Manag. Conf. 2020; pp 1662--1669, DOI: \doi{10.3850/978-981-14-8593-0_3972-cd}\relax
\mciteBstWouldAddEndPuncttrue
\mciteSetBstMidEndSepPunct{\mcitedefaultmidpunct}
{\mcitedefaultendpunct}{\mcitedefaultseppunct}\relax
\EndOfBibitem
\bibitem[Pecht and Gu(2009)Pecht, and Gu]{Pecht2009}
Pecht,~M.; Gu,~J. Physics-of-failure-based prognostics for electronic products. \emph{Trans. Inst. Meas. Control} \textbf{2009}, \emph{31}, 309--322, DOI: \doi{10.1177/0142331208092031}\relax
\mciteBstWouldAddEndPuncttrue
\mciteSetBstMidEndSepPunct{\mcitedefaultmidpunct}
{\mcitedefaultendpunct}{\mcitedefaultseppunct}\relax
\EndOfBibitem
\bibitem[Steenwinckel \latin{et~al.}(2018)Steenwinckel, Heyvaert, De~Paepe, Janssens, Vanden~Hautte, Dimou, De~Turck, Van~Hoecke, and Ongenae]{folio}
Steenwinckel,~B.; Heyvaert,~P.; De~Paepe,~D.; Janssens,~O.; Vanden~Hautte,~S.; Dimou,~A.; De~Turck,~F.; Van~Hoecke,~S.; Ongenae,~F. Towards adaptive anomaly detection and root cause analysis by automated extraction of knowledge from risk analyses. Proc. Semant. Sens. Netw. Workshop (ISWC 2018). 2018; pp 17--31\relax
\mciteBstWouldAddEndPuncttrue
\mciteSetBstMidEndSepPunct{\mcitedefaultmidpunct}
{\mcitedefaultendpunct}{\mcitedefaultseppunct}\relax
\EndOfBibitem
\bibitem[Klein \latin{et~al.}(2025)Klein, Malburg, and Bergmann]{Klein2025}
Klein,~P.; Malburg,~L.; Bergmann,~R. Combining informed data-driven anomaly detection with knowledge graphs for root cause analysis in predictive maintenance. \emph{Eng. Appl. Artif. Intell.} \textbf{2025}, \emph{145}, 110152, DOI: \doi{10.1016/j.engappai.2025.110152}\relax
\mciteBstWouldAddEndPuncttrue
\mciteSetBstMidEndSepPunct{\mcitedefaultmidpunct}
{\mcitedefaultendpunct}{\mcitedefaultseppunct}\relax
\EndOfBibitem
\bibitem[Compton \latin{et~al.}(2012)Compton, Barnaghi, Bermudez, García-Castro, Corcho, Cox, Graybeal, Hauswirth, Henson, Herzog, Huang, Janowicz, Kelsey, {Le Phuoc}, Lefort, Leggieri, Neuhaus, Nikolov, Page, Passant, Sheth, and Taylor]{ssn}
Compton,~M. \latin{et~al.}  The SSN ontology of the W3C semantic sensor network incubator group. \emph{J. Web Semant.} \textbf{2012}, \emph{17}, 25--32, DOI: \doi{https://doi.org/10.1016/j.websem.2012.05.003}\relax
\mciteBstWouldAddEndPuncttrue
\mciteSetBstMidEndSepPunct{\mcitedefaultmidpunct}
{\mcitedefaultendpunct}{\mcitedefaultseppunct}\relax
\EndOfBibitem
\bibitem[Wagner \latin{et~al.}(2023)Wagner, Michels, Schulz, Nair, Rudolph, and Kloft]{Wagner2023}
Wagner,~D.; Michels,~T.; Schulz,~F.~C.; Nair,~A.; Rudolph,~M.; Kloft,~M. TimeSe{AD}: Benchmarking Deep Multivariate Time-Series Anomaly Detection. \emph{Trans. Mach. Learn. Res.} \textbf{2023}, \relax
\mciteBstWouldAddEndPunctfalse
\mciteSetBstMidEndSepPunct{\mcitedefaultmidpunct}
{}{\mcitedefaultseppunct}\relax
\EndOfBibitem
\bibitem[Malhotra \latin{et~al.}()Malhotra, Vig, Shroff, and Agarwal]{malhotra2015a}
Malhotra,~P.; Vig,~L.; Shroff,~G.; Agarwal,~P. Long Short Term Memory Networks for Anomaly Detection in Time Series. Proc. ESANN 23. p 2015–56\relax
\mciteBstWouldAddEndPuncttrue
\mciteSetBstMidEndSepPunct{\mcitedefaultmidpunct}
{\mcitedefaultendpunct}{\mcitedefaultseppunct}\relax
\EndOfBibitem
\bibitem[Kim()]{kim2022a}
Kim,~S. Towards a Rigorous Evaluation of Time-Series Anomaly Detection. Proc. AAAI Conf. Artif. Intell. p 7194–7201, DOI: \doi{10.1609/aaai.v36i7.20680}\relax
\mciteBstWouldAddEndPuncttrue
\mciteSetBstMidEndSepPunct{\mcitedefaultmidpunct}
{\mcitedefaultendpunct}{\mcitedefaultseppunct}\relax
\EndOfBibitem
\bibitem[Soelch \latin{et~al.}(2016)Soelch, Bayer, Ludersdorfer, and van~der Smagt]{soelch2016a}
Soelch,~M.; Bayer,~J.; Ludersdorfer,~M.; van~der Smagt,~P. Variational Inference for On-line Anomaly Detection in High-Dimensional Time Series. ICLR 2016 Workshop. 2016; DOI: \doi{10.48550/arXiv.1602.07109}\relax
\mciteBstWouldAddEndPuncttrue
\mciteSetBstMidEndSepPunct{\mcitedefaultmidpunct}
{\mcitedefaultendpunct}{\mcitedefaultseppunct}\relax
\EndOfBibitem
\bibitem[Shen \latin{et~al.}()Shen, Li, and Kwok]{shen2020a}
Shen,~L.; Li,~Z.; Kwok,~J. Timeseries anomaly detection using temporal hierarchical one-class network. Proc. of the 34th Int. Conf. on Neural Information Processing Systems. 13016–13026, Article 1092.\relax
\mciteBstWouldAddEndPunctfalse
\mciteSetBstMidEndSepPunct{\mcitedefaultmidpunct}
{}{\mcitedefaultseppunct}\relax
\EndOfBibitem
\bibitem[Zhao \latin{et~al.}(2020)Zhao, Wang, Duan, Huang, Cao, Tong, Xu, Bai, Tong, and Zhang]{zhao2020multivariate}
Zhao,~H.; Wang,~Y.; Duan,~J.; Huang,~C.; Cao,~D.; Tong,~Y.; Xu,~B.; Bai,~J.; Tong,~J.; Zhang,~Q. Multivariate time-series anomaly detection via graph attention network. Proc. IEEE Int. Conf. Data Min. (ICDM). 2020; pp 841--850, DOI: \doi{10.1109/ICDM50108.2020.00093}\relax
\mciteBstWouldAddEndPuncttrue
\mciteSetBstMidEndSepPunct{\mcitedefaultmidpunct}
{\mcitedefaultendpunct}{\mcitedefaultseppunct}\relax
\EndOfBibitem
\end{mcitethebibliography}

\providecommand{\latin}[1]{#1}
\makeatletter
\providecommand{\doi}
  {\begingroup\let\do\@makeother\dospecials
  \catcode`\{=1 \catcode`\}=2 \doi@aux}
\providecommand{\doi@aux}[1]{\endgroup\texttt{#1}}
\makeatother
\providecommand*\mcitethebibliography{\thebibliography}
\csname @ifundefined\endcsname{endmcitethebibliography}  {\let\endmcitethebibliography\endthebibliography}{}

\end{document}